\documentclass[10pt,twocolumn,letterpaper]{article}
  
\usepackage{times}
\usepackage{epsfig}
\usepackage{graphicx}
\usepackage{amsmath}
\usepackage{amssymb}

\usepackage[pagenumbers]{cvpr} 

\usepackage[pagebackref,breaklinks,colorlinks]{hyperref}

\def\confName{CVPR}
\def\confYear{2024}

\usepackage{bm}

\usepackage{graphicx}
\usepackage{amsmath}
\usepackage{amssymb}
\usepackage{booktabs}
  
\usepackage{caption}
\usepackage{multirow}

\usepackage{wrapfig}

\usepackage{times}
\usepackage{epsfig}
\usepackage{graphicx}
\usepackage{amsmath}
\usepackage{amssymb}
\usepackage{makecell}
\usepackage{caption}
\usepackage{wrapfig}
\usepackage{booktabs}

\usepackage{pifont}

\usepackage[square, comma, sort&compress, numbers]{natbib}

\usepackage[capitalize]{cleveref}
\crefname{section}{Sec.}{Secs.}
\Crefname{section}{Section}{Sections}
\Crefname{table}{Table}{Tables}
\crefname{table}{Tab.}{Tabs.}

\usepackage{overpic}
\usepackage{enumitem} 
\usepackage{overpic} 
\usepackage{color}

\definecolor{turquoise}{cmyk}{0.65,0,0.1,0.3}
\definecolor{purple}{rgb}{0.65,0,0.65}
\definecolor{dark_green}{rgb}{0, 0.5, 0}
\definecolor{orange}{rgb}{0.8, 0.6, 0.2}
\definecolor{red}{rgb}{0.8, 0.2, 0.2}
\definecolor{darkred}{rgb}{0.6, 0.1, 0.05}
\definecolor{blueish}{rgb}{0.0, 0.3, .6}
\definecolor{light_gray}{rgb}{0.7, 0.7, .7}
\definecolor{pink}{rgb}{1, 0, 1}
\definecolor{greyblue}{rgb}{0.25, 0.25, 1}

\newcommand\cnum[1]{\raisebox{.5pt}{\textcircled{\raisebox{-0.9pt}{#1}}}}

\newcommand*{\appendixdef}{\scalebox{1.2}{\textbf{Appendix.}}}

\usepackage{blindtext}

\renewcommand{\paragraph}[1]{\vspace{1em}\noindent\textbf{#1}.}

\begin{document}

\title{SurMo: Surface-based 4D Motion Modeling for Dynamic Human Rendering}

\author{Tao Hu, ~Fangzhou Hong, ~Ziwei Liu\\
	\vspace{0.12in}
	S-Lab, Nanyang Technological University, Singapore	
}
 

\maketitle
	
\newcommand{\nickname}{SurMo}

\newcommand{\tabCmpZju}{
\begin{table}[t] 
	\small
	\renewcommand{\arraystretch}{1.3}

	\centering
	\caption{Quantitative comparisons on ZJU-MoCap dataset (averaged on all test views and poses on 6 sequences) for novel-view synthesis. To reduce the influence of the background, all scores are calculated from images cropped to 2D bounding boxes. Note that the perception metrics LPIPS \cite{lpip} and FID \cite{Heusel2017GANsTB} capture human judgment better than per-pixel metrics such as SSIM \cite{ssim} or PSNR, as stated in \cite{neuralactor,feanet}. }

    \begin{tabular}{lcccc}
        \hline
        ZJU-S1-6 & LPIPS $\downarrow$ & FID $\downarrow$ & SSIM $\uparrow$ & PSNR $\uparrow$ \\ \hline
        Neural Body~\cite{neuralbody} & .164 & 125.10 & .703 & 21.438 \\ \hline
        Instant-NVR~\cite{instant_nvr} & .202 & 144.99 & .738 & 21.748 \\ \hline
        HumanNeRF~\cite{Weng2022HumanNeRFFR}   & .106 & 88.382  & .792 & 23.624 \\ \hline
        Ours & \textbf{.075} & \textbf{67.725} & \textbf{.833} & \textbf{24.815} \\ \hline
    \end{tabular}
		
  \vspace{-0.12in}
	\label{tab:cmp_zju_sum}
	\end{table}
}

\newcommand{\tabcmpdva}{
\begin{table}[] 
    \small
    \renewcommand{\arraystretch}{1.3}
    \centering
    \caption{{Quantitative comparisons against the 3D pose- and image-driven approach DVA \cite{Remelli2022DrivableVA} and HVTR++~\cite{hvtrpp} on ZJU-MoCap datasets (averaged on all test views and poses) for novel view synthesis. To reduce the influence of the background, all scores are calculated from images cropped to 2D bounding boxes as used in \cite{hvtrpp}. Note that the training and test are conducted at the image resolution of 1024 $\times$ 1024 by following the setup in DVA \cite{Remelli2022DrivableVA}.} For reference, we report the quantitative results of HVTR++ and DVA from the HVTR++ paper.} 
    \begin{tabular}{lcccc}
    \specialrule{.1em}{.1em}{.1em}	
    S386 & LPIPS$\downarrow$ & FID$\downarrow$  & SSIM$\uparrow$          & PSNR$\uparrow$ \\ \hline
    DVA \cite{Remelli2022DrivableVA}   &.146	&117.80	&.791	&26.209 \\
    HVTR++ \cite{hvtrpp} & {.131}	& {84.291}	& {.797}	& {26.517} \\
    Ours & \textbf{.108}	& \textbf{72.556}	& \textbf{.807} &	\textbf{27.164} \\
    \specialrule{.1em}{.1em}{.1em} \specialrule{.1em}{.1em}{.1em}

    S387 & LPIPS$\downarrow$ & FID$\downarrow$ & SSIM$\uparrow$ & PSNR$\uparrow$ \\ \hline	
    DVA \cite{Remelli2022DrivableVA}    & .166	&142.67	&{.791}	&22.474 \\
    HVTR++ \cite{hvtrpp} & {.136} & {101.03}	&.786	& {22.515}	\\ 
    Ours  & \textbf{.112} & \textbf{76.097}	& \textbf{.808}	& \textbf{23.581}	\\ 
    \specialrule{.1em}{.1em}{.1em} 

    \end{tabular}
    \label{tab:cmp_dva}
\end{table}
}

\newcommand{\tabcmparah}{
\begin{table}[] 
    \small
    \renewcommand{\arraystretch}{1.3}
    \centering
    \caption{Quantitative comparisons against ARAH \cite{ARAH:2022:ECCV} for novel view synthesis of training poses and novel poses on ZJU-MoCap datasets (averaged on all test views and poses) for novel view synthesis. To reduce the influence of the background, all scores are calculated from images cropped to 2D bounding boxes.} 
    \begin{tabular}{lcccc}
    \specialrule{.1em}{.1em}{.1em} 

    S377-Train & LPIPS$\downarrow$ & FID$\downarrow$  & SSIM$\uparrow$          & PSNR$\uparrow$ \\ \hline
    ARAH \cite{ARAH:2022:ECCV}   &.096	&83.900	& \textbf{.870}	& 25.176 \\
    Ours & \textbf{.069}	& \textbf{63.008}	& {.866} & \textbf{25.306} \\ \hline

    S386-Train & LPIPS$\downarrow$ & FID$\downarrow$  & SSIM$\uparrow$          & PSNR$\uparrow$ \\ \hline
    ARAH \cite{ARAH:2022:ECCV}   & .112 & 	99.614	&  \textbf{.808}	&  27.008 \\
    Ours 	& \textbf{.080}	&  \textbf{85.811}	&  .801	& \textbf{27.069} \\ 

    \specialrule{.1em}{.1em}{.1em} \specialrule{.1em}{.1em}{.1em}

    S377-Novel & LPIPS$\downarrow$ & FID$\downarrow$  & SSIM$\uparrow$          & PSNR$\uparrow$ \\ \hline
    ARAH \cite{ARAH:2022:ECCV}   &.116	&106.46	&\textbf{.821}	& 23.355 \\
    Ours & \textbf{.088}	& \textbf{78.961}	& {.819} &	\textbf{23.594} \\ \hline

    S386-Novel & LPIPS$\downarrow$ & FID$\downarrow$  & SSIM$\uparrow$          & PSNR$\uparrow$ \\ \hline
    ARAH \cite{ARAH:2022:ECCV}   &.150	&114.24	&\textbf{.742}	& \textbf{25.031} \\
    Ours & \textbf{.123}	& \textbf{104.45}	& {.728} &	{24.821} \\ 

    \specialrule{.1em}{.1em}{.1em} 

    \end{tabular}
    \label{tab:cmp_arah}
\end{table}
}

\newcommand{\tabcmpAIST}{
\begin{table}[] 
    \small
    \renewcommand{\arraystretch}{1.3}
    \centering
    \caption{Quantitative comparisons on S13 and S21 sequences from AIST++ datasets \cite{aist}. To reduce the influence of the background, all scores are calculated from images cropped to 2D bounding boxes.} 
    \begin{tabular}{lcccc}
    \specialrule{.1em}{.1em}{.1em}	
    
    S13 & LPIPS$\downarrow$ & FID$\downarrow$  & SSIM$\uparrow$          & PSNR$\uparrow$ \\ \hline
    
    Neural Body \cite{neuralbody}  &.266 & 276.70 &.732 & \textbf{17.649} \\
    Ours  & \textbf{.183}	& \textbf{161.68}	& \textbf{.751}	& {17.488} \\
    \specialrule{.1em}{.1em}{.1em} \specialrule{.1em}{.1em}{.1em}

    S21 & LPIPS$\downarrow$ & FID$\downarrow$ & SSIM$\uparrow$ & PSNR$\uparrow$ \\ \hline	
    Neural Body \cite{neuralbody}  &.296 & 333.03	&.731 & 17.137 \\
    Ours  & \textbf{.205} & \textbf{177.36}	& \textbf{.757} & \textbf{17.334} \\
    \specialrule{.1em}{.1em}{.1em} 

    \end{tabular}
    \label{tab:cmp_aist}
\end{table}
}

\newcommand{\tabcmpMPI}{
\begin{table}[] 
    \small
    \renewcommand{\arraystretch}{1.3}
    \centering
    \caption{Quantitative comparisons on MPII-RDDC datasets \cite{Habermann2021RealtimeDD}. To reduce the influence of the background, all scores are calculated from images cropped to 2D bounding boxes.} 
    \begin{tabular}{lcccc}
    \specialrule{.1em}{.1em}{.1em}	
    Methods & LPIPS$\downarrow$ & FID$\downarrow$  & SSIM$\uparrow$          & PSNR$\uparrow$ \\ \hline
    HumanNeRF \cite{Weng2022HumanNeRFFR}   &.175	&116.53	&.615	&17.443 \\
    Ours  & \textbf{.153}	& \textbf{107.79}	& \textbf{.627}	& \textbf{18.048} \\
    \specialrule{.1em}{.1em}{.1em} 

    \end{tabular}
    \label{tab:cmp_mpi}
\end{table}
}

\newcommand{\tabcmpDynamics}{
\begin{table}[] 
    \small
    \renewcommand{\arraystretch}{1.3}
    \centering
    \caption{Ablation study of dynamics conditioning.} 
    \begin{tabular}{lcccc}
    \specialrule{.1em}{.1em}{.1em}    
    Methods & LPIPS$\downarrow$ & FID$\downarrow$  & SSIM$\uparrow$          & PSNR$\uparrow$ \\ \hline

    Norm. + Velo. \cite{Yoon2022LearningMA} & .093	& 81.900	& .825	& 24.113 \\
    Ours w/ $D_{cond}$ &.085	&73.674	&.834 &	24.908 \\
    Ours w/ $V_{pred}, N_{pred}$ &.060	&50.170	&.869	&26.654    
    \\
    \specialrule{.1em}{.1em}{.1em} 
    \end{tabular}
    \label{tab:cmp_dynamics}
\end{table}
}

\newcommand{\tabcmpSR}{
\begin{table}[] 
    \small
    \renewcommand{\arraystretch}{1.3}
    \centering
    \caption{Ablation study of super-resolution module under different image resolutions and upsampling factors.} 
    \begin{tabular}{lcccc}
    \specialrule{.1em}{.1em}{.1em}    
    Methods & LPIPS$\downarrow$ & FID$\downarrow$  & SSIM$\uparrow$          & PSNR$\uparrow$ \\ \hline
$512^2$, $\times2$	& .060	& 49.714	& .870	& 26.678 \\
$512^2$, $\times4$	& .070	& 56.456	& .854	& 26.166 \\
$1024^2$, $\times2$	& .076	& 54.563	& .862	& 26.063 \\ 
    \specialrule{.1em}{.1em}{.1em} 
    \end{tabular}
    \label{tab:cmp_sr}
\end{table}
}

\newcommand{\tabCmpZjuAll}{
\begin{table*}[t]
	\vspace{0.06in}	
  \small
  \renewcommand{\arraystretch}{1.3}
	\centering
	\caption{Quantitative comparisons with Neural Body \cite{neuralbody}, Instant-NVR \cite{instant_nvr}, HumanNeRF \cite{Weng2022HumanNeRFFR} on ZJU-MoCap. Instant-NVR* and Instant-NVR are trained with 100 and 30 epochs respectively, which generate better results than the official models that were trained with 6 epochs. Qualitative results can be found in the demo video.}
	\vspace{-0.02in}
  \begin{tabular}{lcccccccccccc}
    \specialrule{.1em}{.1em}{.1em}
    & \multicolumn{4}{l}{S313}   & \multicolumn{4}{l}{S315} & \multicolumn{4}{l} {S377} \\ \hline

    Models & LPIPS $\downarrow$ & FID $\downarrow$ & SSIM $\uparrow$ & PSNR $\uparrow$ & LPIPS & FID & SSIM  & PSNR & LPIPS & FID & SSIM  & PSNR \\ \hline
    Neural Body & .152 & 149.43 & .844 & 26.755 & .108 & 112.57 & .855 & 23.340 & .119 & 132.16 & .862 & 25.997 \\ \hline
    Instant-NVR & .199 & 153.46 & .783 & 23.123 & .230 & 175.68 & .716 & 19.066 & .173 & 123.24 & .810 & 22.976 \\ \hline
    Instant-NVR* & .185 & 132.73 & .783 & 23.029 & .186 & 148.43 & .704 & 18.592 & .159 & 119.97 & .806 & 22.884 \\ \hline
    HumanNeRF & .098 & 69.868 & .822 & 24.870 & .084 & 82.412 & .830 & 21.314 & .092 & 79.760 & .804 & 24.651 \\ \hline
    Ours & .060 & 50.170 & .869 & 26.654 & .058 & 59.664 & .868 & 23.125 & .069 & 63.008 & .866 & 25.306 \\ 
    \specialrule{.1em}{.1em}{.1em} \specialrule{.1em}{.1em}{.1em}
		& \multicolumn{4}{l}{S386} & \multicolumn{4}{l}{S387} & \multicolumn{4}{l}{S394} \\ \hline
    Neural Body & .148 & 133.74 & .815 & 27.648 & .215 & 173.33 & .769 & 23.454 & .217 & 169.12 & .803 & 26.467 \\ \hline
    Instant-NVR & .171 & 137.29 & .742 & 24.639 & .237 & 161.94 & .724 & 20.990 & .251 & 159.11 & .725 & 23.111 \\ \hline
    Instant-NVR* & .161 & 135.96 & .736 & 24.591 & .230 & 155.97 & .724 & 21.070 & .247 & 155.41 & .727 & 23.244 \\ \hline
    HumanNeRF & .105 & 100.43 & .763 & 26.590 & .129 & 96.722 & .762 & 22.452 & .119 & 97.947 & .766 & 24.643 \\ \hline
    Ours & .080 & 85.811 & .801 & 27.069 & .084 & 71.216 & .810 & 23.735 & .095 & 78.949 & .787 & 25.237 \\ 
    \specialrule{.1em}{.1em}{.1em} 

    \end{tabular}
    \vspace{-0.06in}
    \label{tab:cmp_zju_all}
  \end{table*}
}

\newcommand{\tabAbMotion}{
\begin{table}[t] 
	\small
	\renewcommand{\arraystretch}{1.3}
	\centering
	\caption{Ablation study of motion conditioning and learning. $P_{cond}$ and $D_{cond}$ denote the conditioning of pose and dynamics, $V_{pred}$ and $N_{pred}$ denote the prediction of surface velocity and surface normal in motion learning.} 
   \vspace{-0.12in}
    \begin{tabular}{lcccc}
        \hline
        S313 & LPIPS $\downarrow$ & FID $\downarrow$ & SSIM $\uparrow$ & PSNR $\uparrow$ \\ \hline

        $P_{cond}$ & .120 & 104.83 & .793 & 22.781 \\ \hline
        $+ \; D_{cond}$ & .085 & 73.674 & .834 & 24.908 \\ \hline
        $+ \; V_{pred}$ & .069 & 62.092 & .856 & 25.845 \\ \hline
        $+ \; N_{pred}$ & .060 & 50.170 & .869 & 26.654 \\ \hline

    \end{tabular}		
  \vspace{-0.12in}
	\label{tab:AbMotion}
	\end{table}
}

\newcommand{\tabAbSupMotion}{
\begin{table}[h]
  \small
  \begin{center}
      \caption{Ablation study of motion prediction and training views.}
      \begin{tabular}{lcccc}
        \specialrule{.1em}{.1em}{.1em} 
          S313 & LPIPS $\downarrow$ & FID $\downarrow$ & SSIM$\uparrow$ & PSNR$\uparrow$\\ \hline
          ${w/o~Pred}$       & .085  & 73.674 & {.834} & {24.908} \\
          $Pred_{\color{red}{t}}$    & .073  & 60.942 & .848 & 25.537 \\            
          $Pred_{\color{red}{t+1}}$    & \textbf{.060}  & \textbf{50.170} & \textbf{.869} & \textbf{26.654} \\ \hline
          $Pred_{t+1}(1~view)$    & {.126}  & {112.19} & {.788} & {22.830} \\  \specialrule{.1em}{.1em}{.1em} \specialrule{.1em}{.1em}{.1em}
          S387 & LPIPS $\downarrow$ & FID $\downarrow$ & SSIM$\uparrow$ & PSNR$\uparrow$\\ \hline
          ${w/o~Pred}$ & .115 &	93.688 & .761 &	22.152	\\
          $Pred_{\color{red}{t}}$   & .096 & 83.825 & .790 &	23.083	\\
          $Pred_{\color{red}{t+1}}$ & \textbf{.084} &	\textbf{71.216} & \textbf{.810} & \textbf{23.735}	\\ \hline
          $Pred_{t+1}(1~view)$   & .151 & 128.18 & .729	& 21.093 \\ 
          \specialrule{.1em}{.1em}{.1em}            
      \end{tabular}
      \label{tab:absupMotion}
  \end{center}    
\end{table}
}
\newcommand{\figTeaser}{
\begin{figure}[t]
	\begin{center}
		\includegraphics[width=\linewidth]{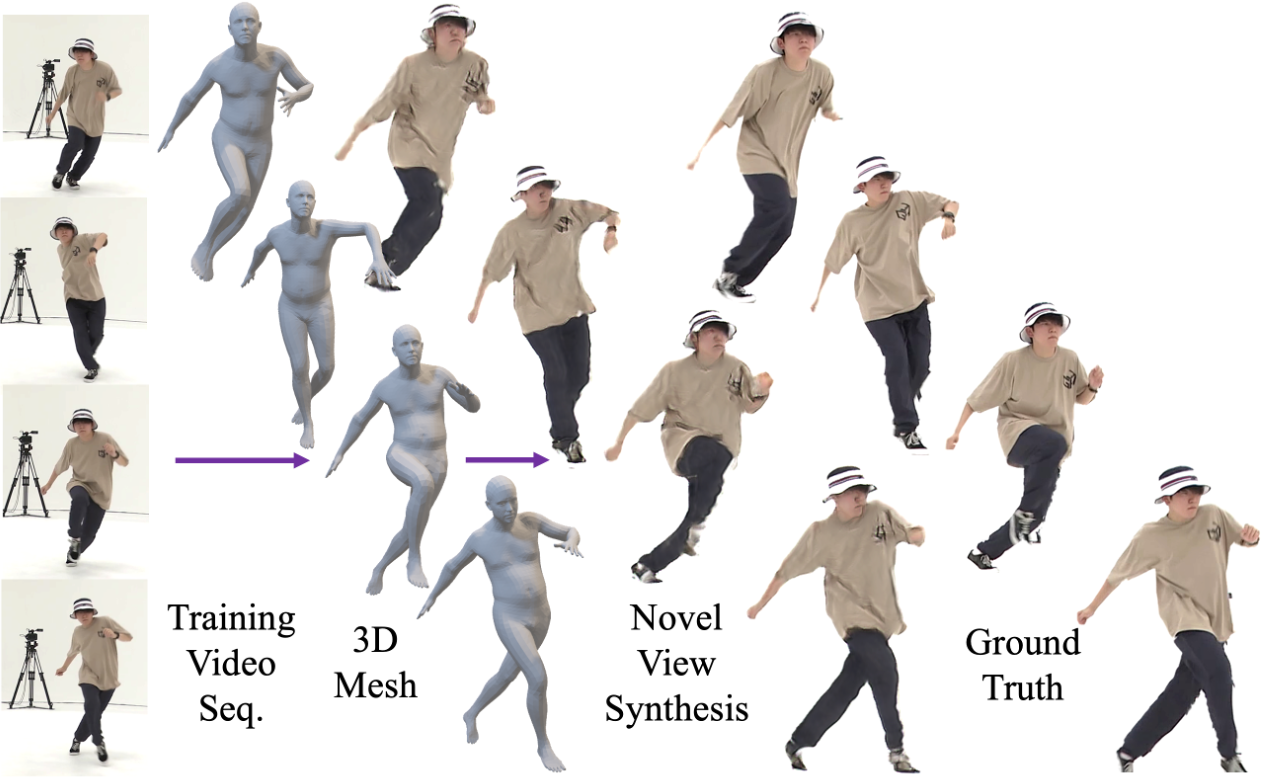}
	\end{center}
	\vspace{-20pt}
	\caption{Given several sparse multi-view video sequences with estimated 3D body meshes, \nickname{} synthesizes subject-specific appearance. We specifically focus on the synthesis of plausible time-varying appearances by learning an effective 4D motion representation.}
	\label{fig:teaser}
	\vspace{-0.12in}
\end{figure}
}

\newcommand{\figFramework}{
\begin{figure*}[t]
	\begin{center}
		\includegraphics[width=\linewidth]{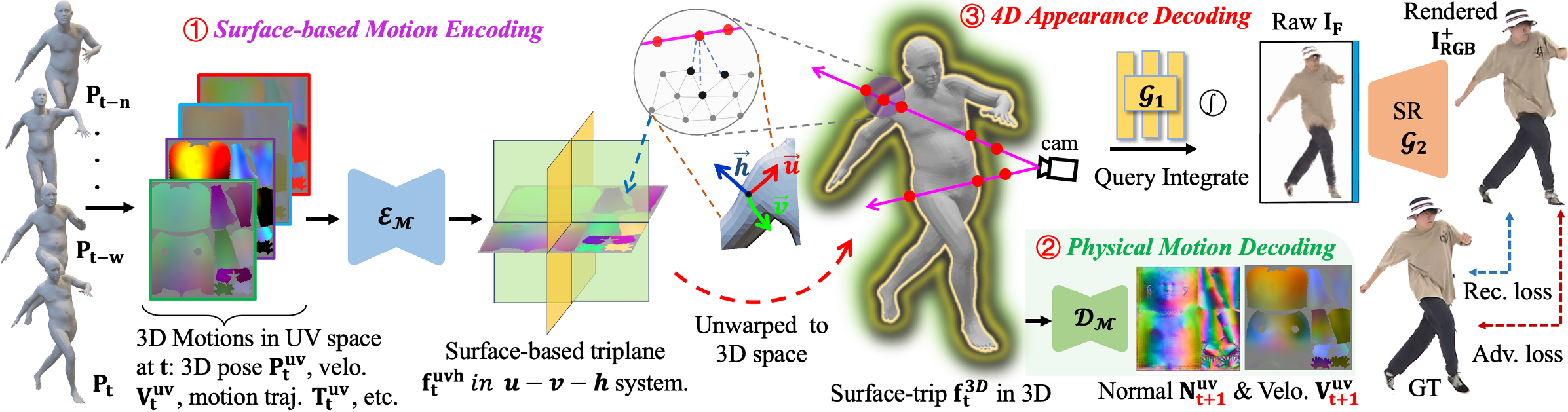}
	\end{center}
	\vspace{-0.14in}
	\caption{Framework overview. Given a set of time-varying 3D body meshes \{$\mathbf{P_t}$, ..., $\mathbf{P_t-n}$\} obtained from training video sequences, we aim to synthesize high-fidelity appearances of a clothed human in motion via a feature encoder-decoder framework: \textit{Motion Encoding}, and joint \textit{Motion} and \textit{Appearance Decoding}. \textbf{1)} We take as input an expressive 4D motion representation extracted from the mesh sequences including 3D pose, 3D velocity at time \textit{t}, and motion trajectory over the past $w$ timesteps that encode both spatial and temporal relations of the motion sequence, which are projected to the spatially aligned UV surface space. A motion encoder $\mathcal{E_M}$ is employed to lift the 2D UV-aligned features to a 3D surface-based triplane $\mathbf{f^{uvh}_t}$ in an UV-plus-height space with a signed distance height to model temporal clothing offsets.   \textbf{2)} A motion decoder $\mathcal{D_M}$ is designed to encourage physical motion learning in training by decoding the triplane features $\mathbf{f^{uvh}_t}$ to predict the motion at the next timestep \textit{t + 1}, \ie spatial derivatives surface normal $\mathbf{N^{uv}_{t+1}}$ and temporal derivatives surface velocity $\mathbf{V^{uv}_{t+1}}$ in UV space. \textbf{3)} Finally, given a target camera view, the triplane $\mathbf{f^{uvh}_t}$ is rendered into high-quality images by a volumetric surface-conditioned renderer including volumetric low-resolution rendering by $\mathcal{G}_1$ and an efficient geometry-aware super-resolution by $\mathcal{G}_2$.}
    \label{fig:framework}
	\vspace{-0.12in}
\end{figure*}
}

\newcommand{\figCmpMotionMain}{
\begin{figure*}[t]
	\begin{center}
		\includegraphics[width=\linewidth]{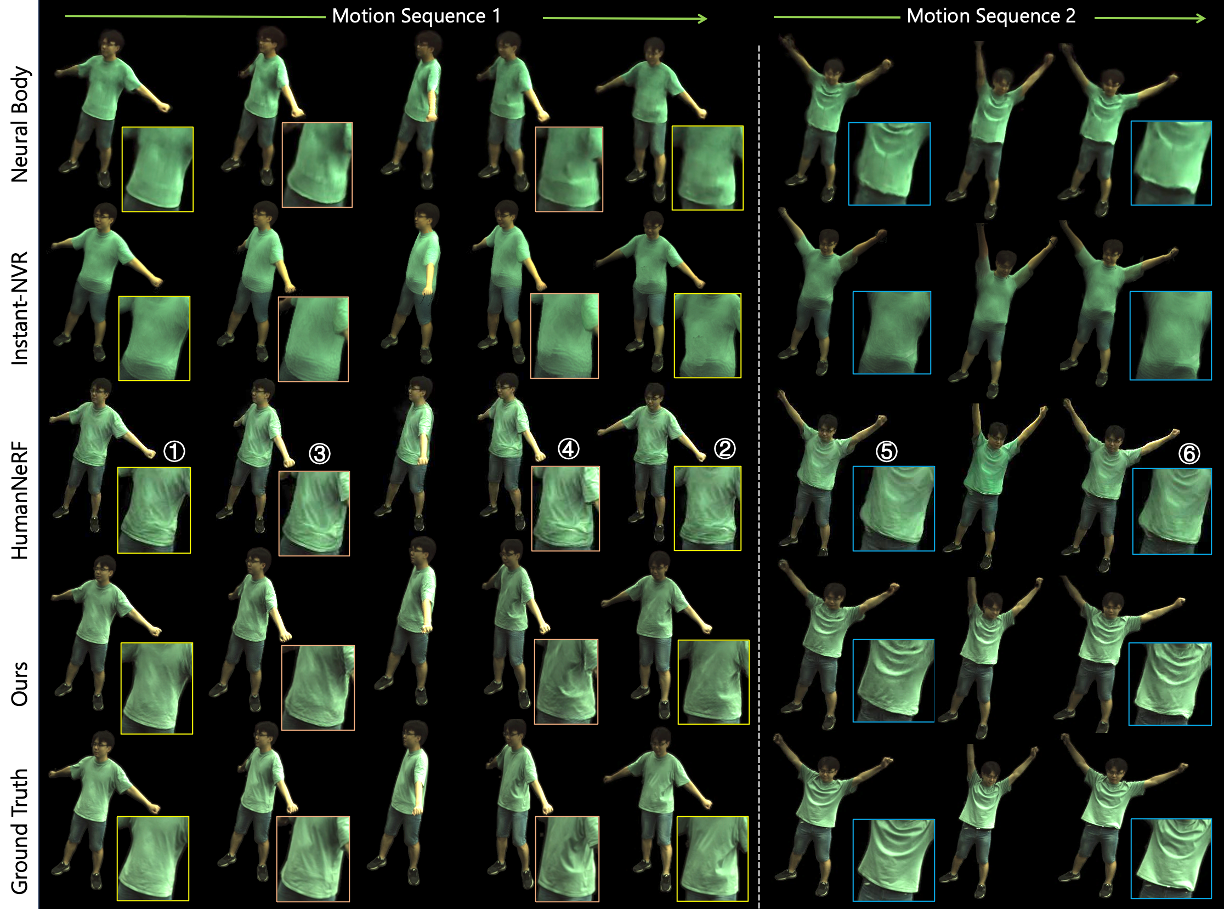}
	\end{center}
	\vspace{-0.14in}
	\caption{Qualitative comparisons on novel view synthesis on the subject S313 of ZJU-MoCap dataset. Two motion sequences S1 (swing arms left to right) and S2 (raise and lower arms) are shown. We specifically focus on the synthesis of time-varying appearances (especially T-shirt wrinkles), by evaluating the rendering results under similar poses yet with different movement directions, which are marked in the same color, such as the pairs of \cnum{1}\cnum{2}, \cnum{3}\cnum{4}, and \cnum{5}\cnum{6}. Our method synthesizes high-fidelity time-varying appearances, whereas SOTA HumanNeRF generates almost the same cloth wrinkles.}
	\label{fig:cmp_313}
\end{figure*}
}

\newcommand{\figCmpMotionZ}{
\begin{figure*}[t]
	\begin{center}
		\includegraphics[width=\linewidth]{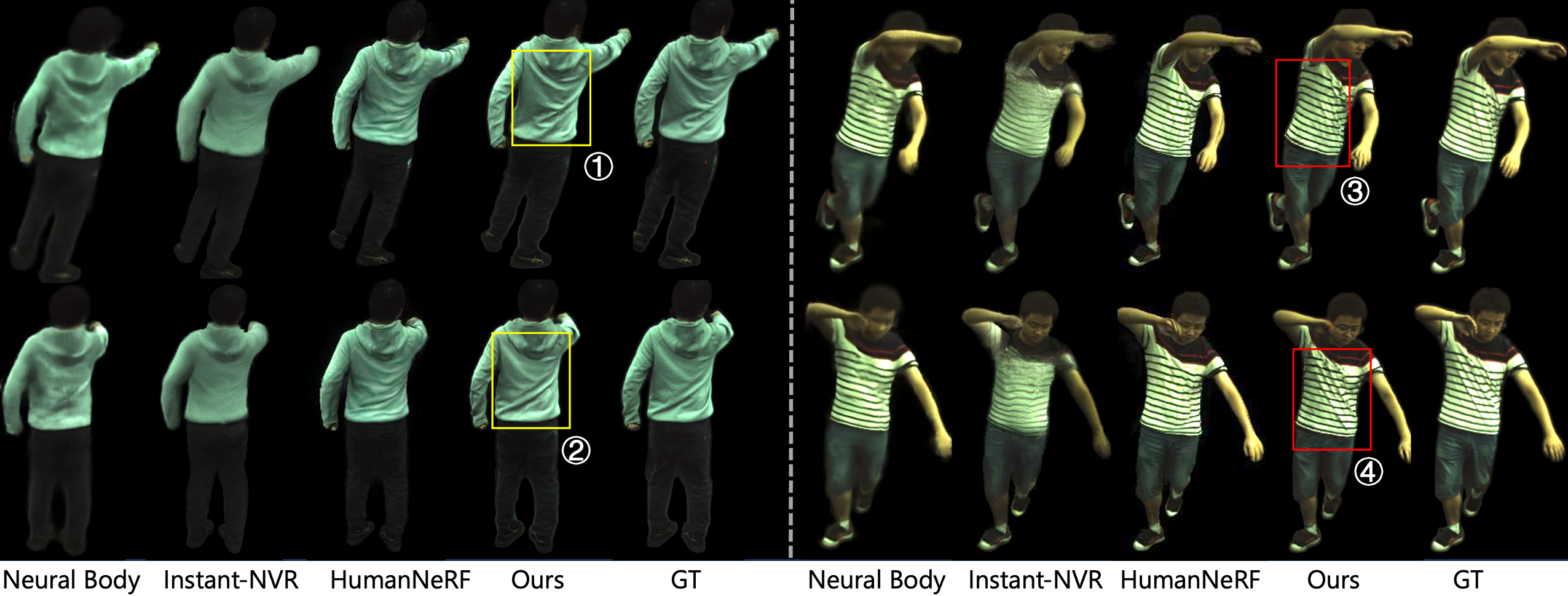}
	\end{center}
	\vspace{-0.14in}
	\caption{Qualitative comparisons on novel view synthesis on the subject S387, S315 of ZJU-MoCap dataset. Row 1 and 2 show similar poses occurring at different timesteps (not consecutive frames). The results indicate that our method synthesizes   
  time-varying appearances while other methods mainly generate pose-dependent appearances.}
	\label{fig:cmp_zju_2}
\end{figure*}
}

\newcommand{\figMpiLighting}{
\begin{figure*}[t]
	\begin{center}
		\includegraphics[width=\linewidth]{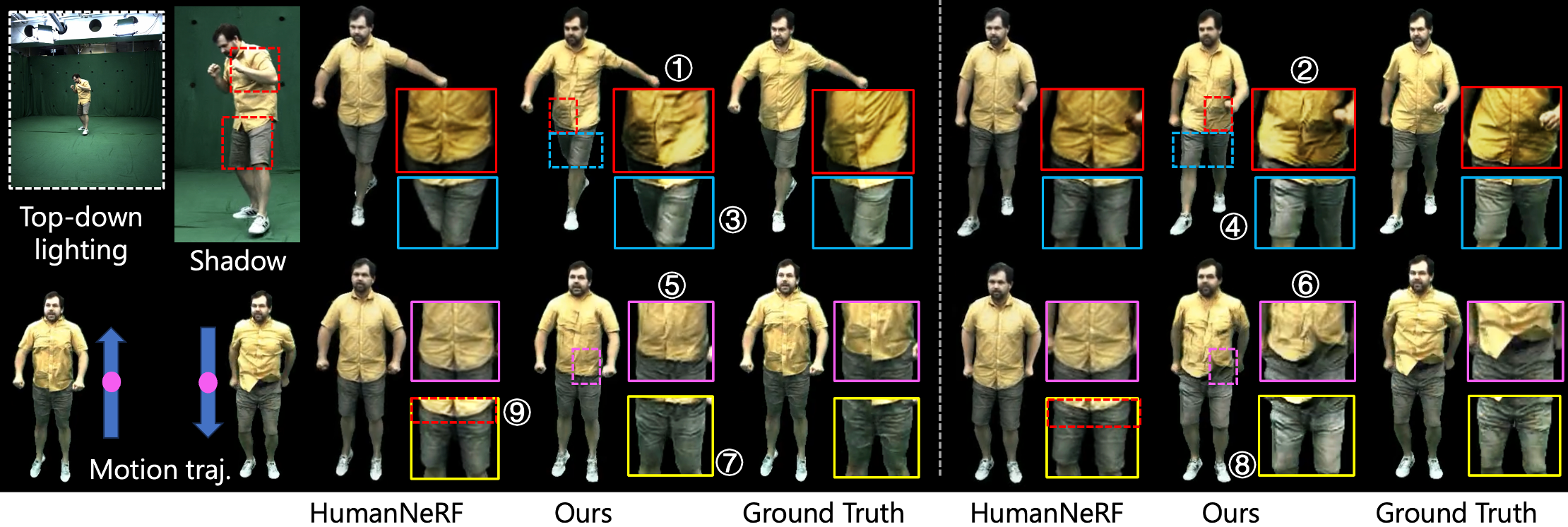}
	\end{center}
	\vspace{-0.12in}
	\caption{Novel view synthesis of time-varying appearances with both pose and lighting conditioning on MPII-RDDC dataset. The sequence is captured in a studio with top-down lighting that casts shadows on the human performer due to self-occlusion. In Row 1, we specifically focus on  synthesizing time-varying shadows (\eg, \cnum{1} vs. \cnum{2}, and \cnum{3} vs. \cnum{4}) for different poses with different self-occlusions. In Row 2, we evaluate the synthesis of: 1) time-varying appearances for similar poses occurring in a jump-up-and-down motion sequence, \eg, \cnum{5} vs. \cnum{6}, 2) shadows \cnum{7} vs. \cnum{8}, and 3) clothing offsets \cnum{5} vs. \cnum{6}. } 
	\label{fig:lighting}
\end{figure*}
}

\newcommand{\figAbTrip}{
\begin{table*}[t]
	\begin{minipage}{.36\linewidth}
		\begin{center} 
			\includegraphics[width=\linewidth]{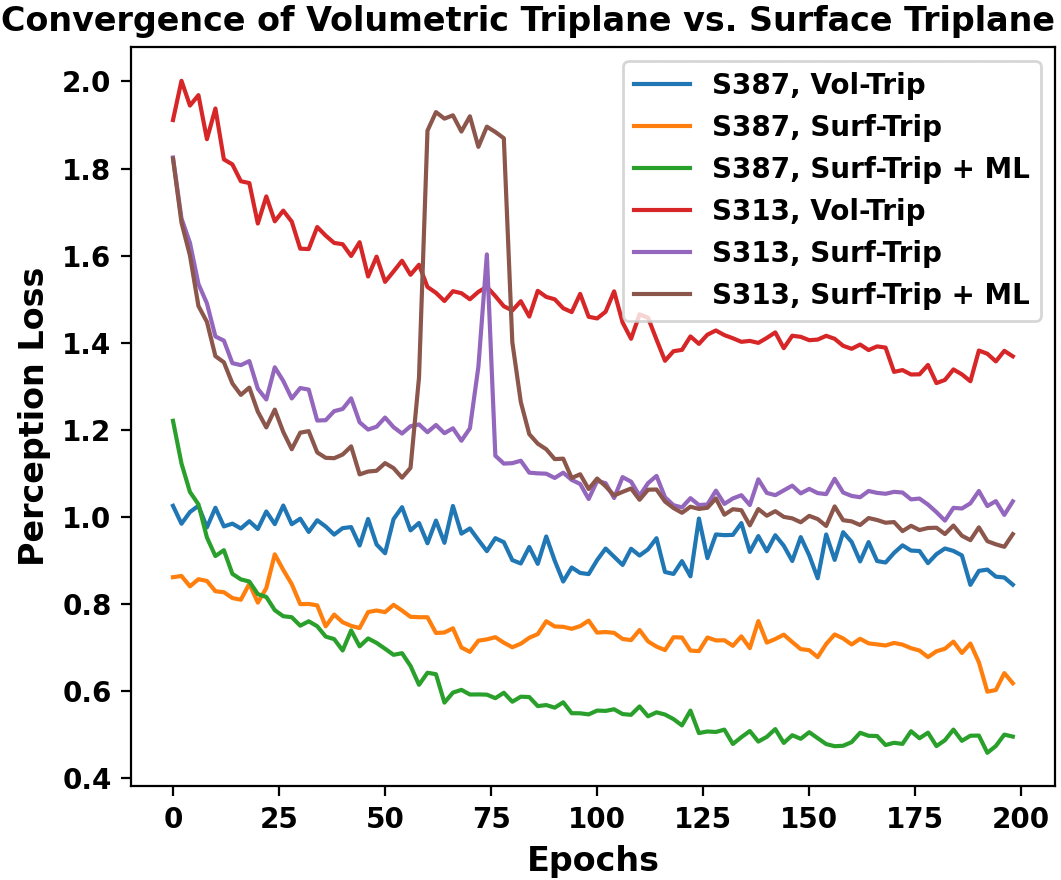}
		\end{center}
	\end{minipage}
	\hfill
	\begin{minipage}{.61\linewidth}
		\begin{center}
			\includegraphics[width=\linewidth]{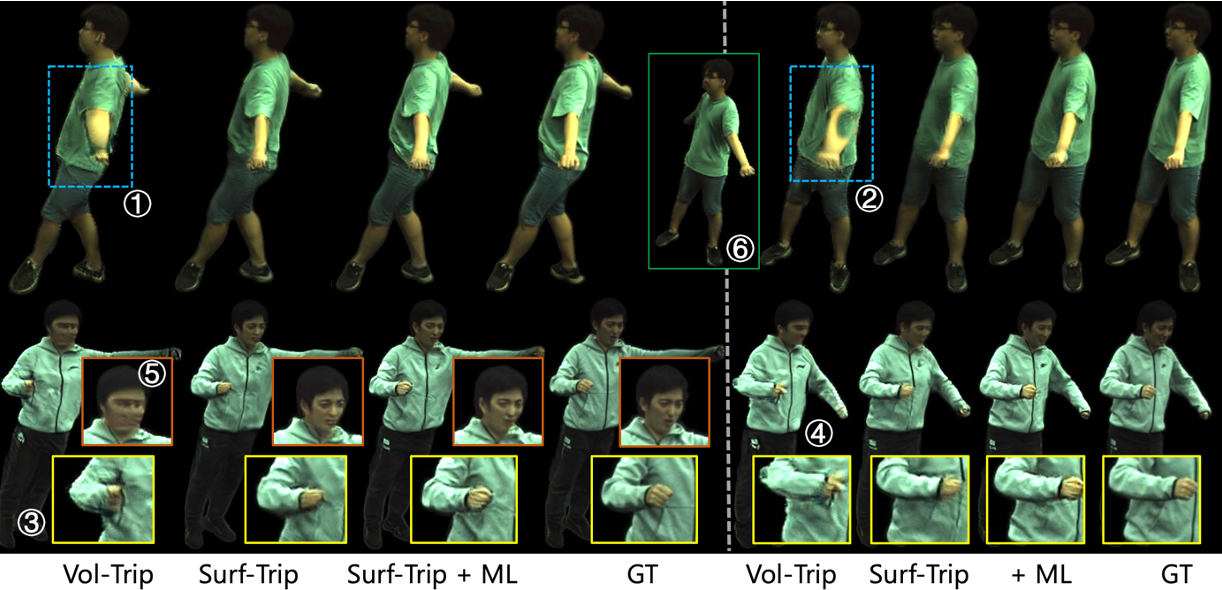}
		\end{center}
	\end{minipage}
 	\vspace{-0.1in}
  \captionof{figure}{Ablation study of 3D volumetric triplane (Vol-Trip) vs. surface-based triplane (Surf-Trip) for human modeling on subject S313 and S387 from ZJU-MoCap. We focus on the convergence in training (left), and self-occlusion rendering (right). The convergence of S313 is shifted for visualization purposes. Vol-Trip is not effective in handling self-occlusion, \eg, \cnum{1}\cnum{2}\cnum{3}\cnum{4} though it performs well in another viewpoint without self-occlusions \cnum{6}. In addition, Vol-Trip cannot synthesize high-quality details for face.}
    \label{fig:ab_trip}
\end{table*}
}

\newcommand{\figAist}{
\begin{figure}[]
	\begin{center}
		\includegraphics[width=\linewidth]{img/cmp_aist.png}
	\end{center}
	\vspace{-0.12in}
	\caption{Novel view synthesis of fast motions on AIST++. We focus on the synthesis of time-varying wrinkles of the T-shirt for different motions.}
	\label{fig:aist}
	\vspace{-0.04in}
\end{figure}
}

\newcommand{\figAistb}{
\begin{figure}[]
	\begin{center}
		\includegraphics[width=\linewidth]{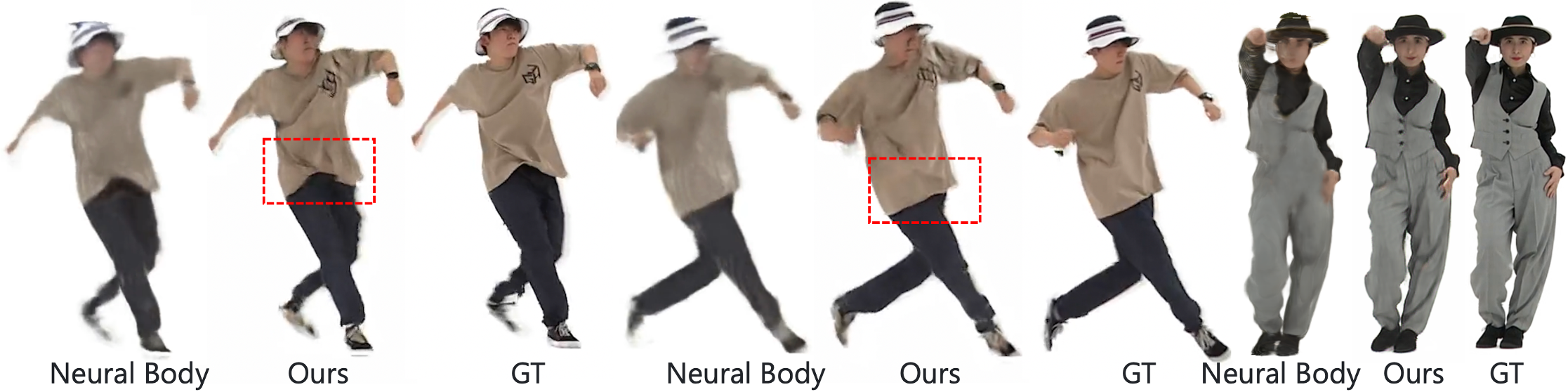}
	\end{center}
	\vspace{-0.12in}
	\caption{Novel view synthesis of fast motions on AIST++.}
	\label{fig:aist}
	\vspace{-0.04in}
\end{figure}
}

\newcommand{\figAistbCam}{
\begin{figure}[]
	\begin{center}
		\includegraphics[width=\linewidth]{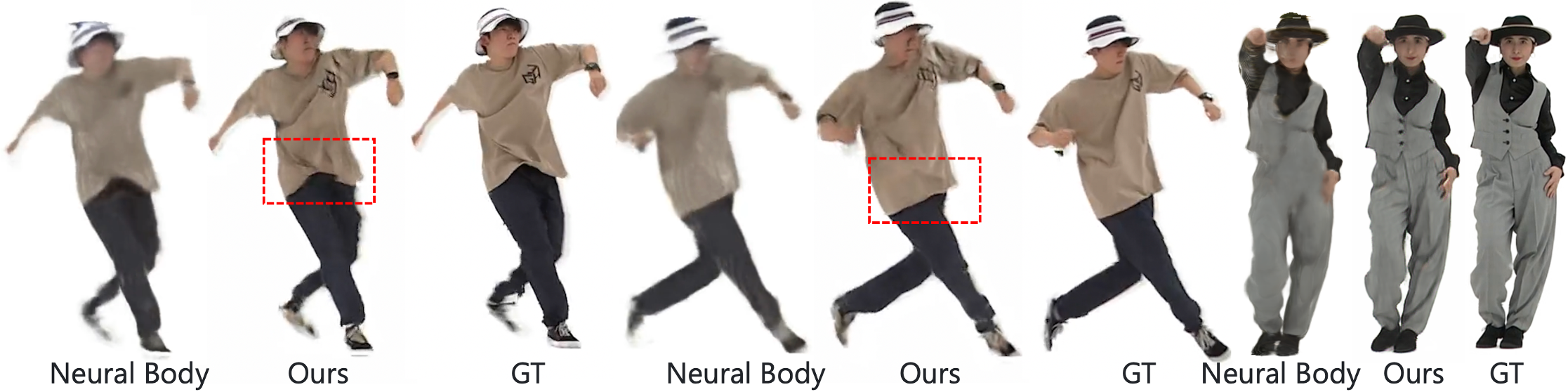}
	\end{center}
	\vspace{-0.12in}
	\caption{Novel view synthesis of fast motions on AIST++.}
	\label{fig:aist}
	\vspace{-0.04in}
\end{figure}
}

\newcommand{\figMotionCond}{
\begin{figure}[]
	\begin{center}
		\includegraphics[width=\linewidth]{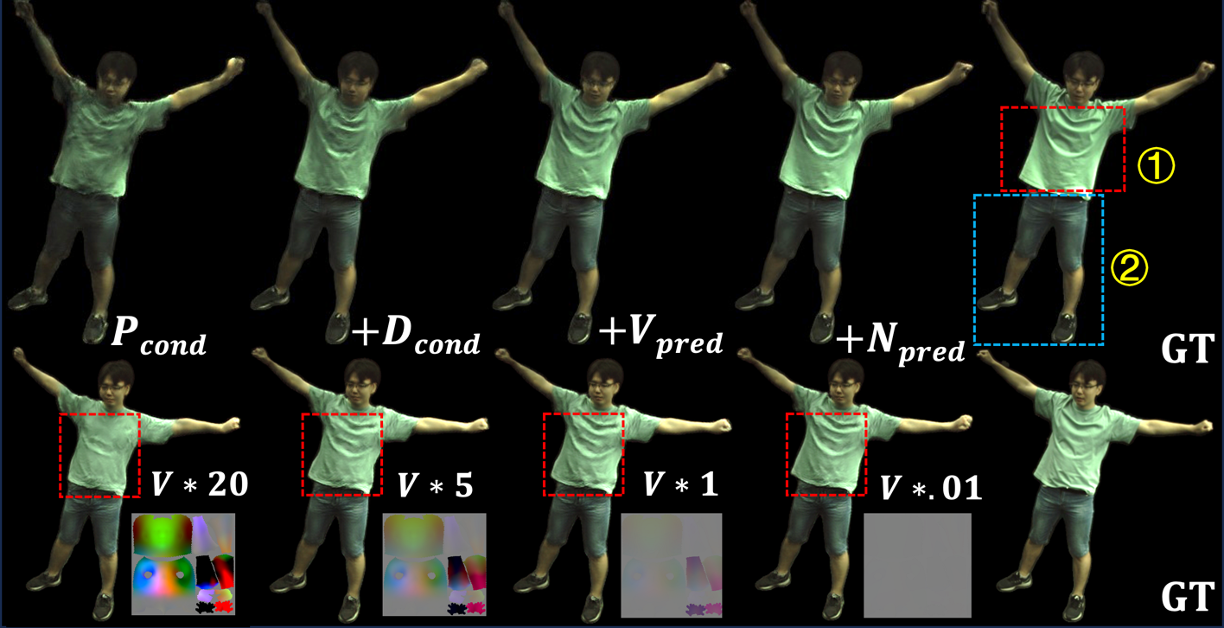}
	\end{center}
	\vspace{-0.12in}
	\caption{Ablation study of motion conditioning and learning. We focus on the effect of motion conditioning and learning in Row 1, and whether our method learns to decouple the static pose and dynamics (\eg, velocity) from the motion conditioning.}
	\label{fig:ab_motion}
	\vspace{-0.12in}
\end{figure}
}

\begin{abstract}

Dynamic human rendering from video sequences has achieved remarkable progress by formulating the rendering as a mapping from static poses to human images. However, existing methods focus on the human appearance reconstruction of every single frame while the temporal motion relations are not fully explored. In this paper, we propose a new 4D motion modeling paradigm, \textbf{SurMo}, that jointly models the temporal dynamics and human appearances in a unified framework with three key designs:
\textbf{1) Surface-based motion encoding} that models 4D human motions with an efficient compact surface-based triplane. It encodes both spatial and temporal motion relations on the dense surface manifold of a statistical body template, which inherits body topology priors for generalizable novel view synthesis with sparse training observations. \textbf{2) Physical motion decoding} that is designed to encourage physical motion learning by decoding the motion triplane features at timestep $t$ to predict both spatial derivatives and temporal derivatives at the next timestep $t+1$ in the training stage.  \textbf{3) 4D appearance decoding} that renders the motion triplanes into images by an efficient volumetric surface-conditioned renderer that focuses on the rendering of body surfaces with motion learning conditioning. Extensive experiments validate the state-of-the-art performance of our new paradigm and illustrate the expressiveness of surface-based motion triplanes for rendering high-fidelity view-consistent humans with fast motions and even motion-dependent shadows. Our project page is at: \href{https://taohuumd.github.io/projects/SurMo/}{https://taohuumd.github.io/projects/SurMo}.

\end{abstract}
\section{Introduction}
\label{sec:intro}

\figTeaser
  
\noindent Creating volumetric videos of human actors is required in many applications such as AR/VR, telepresence, video game and film character creation. Recent neural rendering methods \cite{Tewari2020StateOT,neuralbody,neuralactor,Weng2022HumanNeRFFR} have made great progress in generating free-viewpoint videos of humans from several sparse multi-view videos, which are simple yet effective compared with traditional graphics approaches \cite{Borshukov2003UniversalCI,Carranza2003FreeviewpointVO,Xu2011VideobasedCC}. At the core of the technique is to model pose- and time-varying appearances of dynamic humans. The motions of dressed humans are often expressed as the movement of body and a sequence of natural secondary motion of clothes, \eg, dynamic movements of the T-shirt induced by dancing in Fig. \ref{fig:teaser}. The secondary motion of clothes arises from intricate physical interactions with the body, typically changing over time, which leads to challenges for plausible rendering of dynamic humans. 

Most existing methods \cite{neuralbody,neuralactor,Weng2022HumanNeRFFR} are conditioned on static poses and use a pose-guided generator to synthesize time-varying appearances. However, the appearance of clothed humans undergoes complex geometric transformations induced not only by the static pose but also its dynamics, whereas the modeling of dynamics is often ignored. In addition, these methods are often focused on the 2D appearance reconstruction of every single frame while the temporal motion relations are not fully explored. Due to these issues, they fail to render plausible secondary motions of clothes, \ie, generating the same appearances for fast and slow motions. A key challenge of learning temporal dynamics lies in the requirement of a tremendous amount of training observations to construct a 4D motion volume.

To solve these issues, we propose a new paradigm to learn time-varying appearances of the secondary motion from just several sparse viewpoint video sequences, which is achieved by jointly modeling the temporal motion dynamics and human appearances in a unified rendering framework based on an efficient surface-based motion representation. At the core of the paradigm is a feature encoder-decoder framework with three key components: \textbf{1)} surface-based motion encoding; \textbf{2)} physical motion decoding; and \textbf{3)} 4D appearance decoding. 

Firstly, in contrast to existing pose-guided methods \cite{neuralactor,neuralbody,Peng2021AnimatableNR,Weng2022HumanNeRFFR} that focus on static poses as a conditional variable, we extract an expressive 4D motion input from the 3D body mesh sequences obtained from training video as our input, which includes both a static pose represented by a spatial 3D mesh and its temporal dynamics. Furthermore, we notice that the non-rigid deformations of garments typically occur around the body surface instead of in a 3D volume, and hence we propose to model human motions on the body surface by projecting the extracted spatial-temporal 4D motion input to the dense 2D surface UV manifold of a clothless body template (\eg, SMPL \cite{smpl}). To model temporal clothing offsets, a motion encoder is employed to lift the clothless motion features into a motion triplane that encodes both spatial and temporal motion relations in a compact 3D triplane with time-varying dynamics conditioned. The triplane is defined in the surface $\mathbf{u-v-h}$ coordinate system, with $\mathbf{u-v}$ to represent the motion of the clothless body template, and an extensional coordinate $\mathbf{h}$ to represent the secondary motion of clothes, \ie,  the temporal clothing offsets are parameterized by a signed distance to the body surface. In this way, 4D motions can be effectively represented by a surface-based triplane.

Secondly, we propose to physically model the spatial and temporal motion dynamics in the rendering network. Specifically, with the surface-based triplane conditioned at time $t$, a motion decoder is employed to decode the motion triplanes to predict the motion at the next timestep $t+1$, \ie, spatial derivative of the motion physically corresponding to surface normal map and temporal derivatives corresponding to surface velocity. We illustrate the physical motion learning significantly improves the rendering quality.  
   
Thirdly, the motion triplanes are decoded into high-quality images at two stages: a volumetric surface-conditioned renderer that is focused on the rendering around human body surface and filters the query points far from the body surface for efficient volumetric rendering, and a geometry-aware super-resolution module for efficient high-quality image synthesis.

We conduct a systematical analysis of how human appearances are affected by temporal dynamics, and it was observed that some baseline methods mainly generate pose-dependent appearances instead of time-varying appearances for free-view video generations. In addition, quantitative and qualitative experiments are performed on three datasets with a total of 9 subject sequences, including ZJU-MoCap \cite{neuralbody}, AIST++ \cite{aist}, and MPII-RDDC \cite{Habermann2021RealtimeDD}, which validate the effectiveness of \nickname{} in different scenarios.

In summary, our contributions are:
   
\textbf{1)} A new paradigm for learning dynamic humans from videos that jointly models temporal motions and human appearances in a unified framework, and one of the early works that systematically analyzes how human appearances are affected by temporal dynamics.

\textbf{2)} An efficient surface-based triplane that encodes both spatial and temporal motion relations for expressive 4D motion modeling.

\textbf{3)} We achieve state-of-the-art results and show that our new paradigm is capable of learning high-fidelity appearances from fast motion sequences (\eg., AIST++ dance videos) or synthesizing motion-dependent shadows in challenging scenarios.

\section{Related Work}

\noindent Our method is closely related to many sub-fields of visual computing,  and 
below we discuss a set of the work. 
 
\noindent \textbf{3D Shape Representations.} To capture detailed shapes of 3D objects, most recent papers utilize implicit functions \cite{Mescheder2019OccupancyNL,Michalkiewicz2019DeepLS,Chen2019LearningIF,Park2019DeepSDFLC,Saito2019PIFuPI,Saito2020PIFuHDMP,Huang2020ARCHAR,Saito2021SCANimateWS,Mihajlovi2021LEAPLA,Wang2021MetaAvatarLA,Palafox2021NPMsNP,Tiwari2021NeuralGIFNG,Zheng2021PaMIRPM,Jeruzalski2020NASANA,Zheng2021DeepMultiCapPC} or point clouds \cite{scale,pop,densepointclouds} due to their topological flexibility. These methods aim to learn geometry from 3D datasets, whereas we synthesize human images of novel poses only from 2D RGB training images.  

\noindent \textbf{Rendering Humans by 2D GANs.} Some existing approaches render human avatars by neural image translation, \ie, they utilize GAN \cite{gans} networks to learn a mapping from poses (given in the form of renderings of a skeleton~\cite{edn,SiaroSLS2017,Pumarola_2018_CVPR,KratzHPV2017,zhu2019progressive,vid2vid}, dense mesh~\cite{Liu2019,vid2vid,liu2020NeuralHumanRendering,feanet,Neverova2018,Grigorev2019CoordinateBasedTI} or joint position heatmaps~\cite{MaSJSTV2017,Aberman2019DeepVP,Ma18}) to human images \cite{vid2vid,edn}. To improve temporal stability, some methods \cite{dnr,smplpix,egorend,anr} propose to utilize the SMPL \cite{smpl} priors for pose-guided generations. However, these methods do not reconstruct geometry explicitly and cannot handle self-occlusions effectively. 

\noindent \textbf{Rendering Humans by 3D-aware Renderer.} For stable view synthesis, recent papers \cite{neuralbody,neuralactor,Peng2021AnimatableNR,narf,anerf,Chen2021AnimatableNR} propose to unify geometry reconstruction with view synthesis by volume rendering, which, however, is computationally heavy. To solve this issue, some recent 3D-GAN papers \cite{Niemeyer2021GIRAFFERS,Zhou2021CIPS3DA3,Gu2021StyleNeRFAS,OrEl2021StyleSDFH3,Hong2021HeadNeRFAR,Chan2021EfficientG3,hvtr,hvtrpp} propose a hybrid rendering strategy for efficient geometry-aware rendering, that is, render low-resolution volumetric features for geometry learning, and employ a super-resolution module for high-resolution image synthesis. We adopt this strategy for efficient rendering, whereas ours is distinguished by rendering articulated humans with 4D motion modeling.

\noindent \textbf{UV-based Pose Representation.} Some methods \cite{scale,pop,Remelli2022DrivableVA,hvtr,hvtrpp,Yoon2022LearningMA} propose to project 3D posed meshes into a 2D positional map for pose encoding, which can be used for different downstream tasks, such as 3D reconstruction and novel view synthesis. \cite{scale,pop} rely on 3D supervision to learn 3D reconstructions, and they do not take as input the dynamics. To improve the rendering quality, \cite{Remelli2022DrivableVA, hvtrpp} proposes to utilize additional driving views as input for faithful rendering in telepresence applications. However, they do not explore how to learn temporal dynamics from pose sequences. By taking as input a 3D normal and velocity map, \cite{Yoon2022LearningMA} encodes motion dynamics in their rendering network, whereas they do explicitly learn motions, \ie, predicting the motion status at the next timestep, and besides, \cite{Yoon2022LearningMA} is a 2D rendering method, and they do not explicitly learn motions.
\section{Methodology}
\figFramework


\subsection{Problem Setup}


\noindent Given a sparse multi-view video of a clothed human in motion and corresponding 3D pose estimations $\{\mathbf{P_0}, ..., \mathbf{P_t}\}$, our goal is to synthesize time-varying appearances of the individual under novel views. Existing methods \cite{neuralactor,Weng2022HumanNeRFFR,Peng2021AnimatableNR,neuralbody} formulate the problem of human rendering as learning a representation via a feature encoder-decoder framework: 

\begin{equation}
\label{eq:form_other}
\begin{aligned}
 \mathbf{f_t} & = \mathcal{E_P}(\mathbf{P_t}, \mathbf{z_t})  \\
 \mathbf{I_t} & = \mathcal{D_R}(\mathbf{f_t})
\end{aligned}
\end{equation}
where a pose encoder $\mathcal{E_P}$ takes as input a representation of posed body $\mathbf{P_t}$ (\eg, 2D or 3D keypoints or 3D body mesh vertices) and a timestamp embedding $\mathbf{z_t}$ at time $t$, and outputs intermediate pose-dependent features $\mathbf{f_t}$ that can be rendered by a decoder $\mathcal{D_R}$ to reconstruct the appearance images $\mathbf{I_t}$ of the corresponding pose at time $t$. The time embedding $\mathbf{z_t}$ often serves as a residual that is updated passively in backpropagation by image reconstruction loss. 

However, there are two issues with the above well-adopted paradigm. First, the appearance of clothed humans undergoes complex geometric transformations induced not only by the static pose $\mathbf{P_t}$ but also its dynamics, whereas the modeling of dynamics is ignored in E.q. \ref{eq:form_other}, and the residual $\mathbf{z_t}$ cannot expressively model physical dynamics neither. Second, existing methods focus on the per-image reconstruction while ignoring the temporal relations of a motion sequence in training, \ie, sampling input pose $\mathbf{P_t}$ and timestamp $\mathbf{z_t}$ individually, and supervising image reconstructions frame-by-frame, partly because defining temporal supervision in the 2D image space is challenging, especially for articulated humans that often suffer from misalignment problems due to pose estimation errors. 

To solve these issues, we propose a new paradigm for learning view synthesis of dynamic humans from video sequences, which is formulated as:
\begin{equation}
\label{eq:form_ours}
\begin{aligned}
 \mathbf{f_t} & = \mathcal{E_M}(\mathbf{P_t}, \mathbf{D_t})  \\
 \frac{\partial \mathbf{P_{t+1}}}{\partial \mathbf{x}}, \frac{\partial \mathbf{P_{t+1}}}{\partial \mathbf{t}} & = \mathcal{D_M}(\mathbf{f_t}) \\
 \mathbf{I_t} & = \mathcal{D_R}(\mathbf{f_t})
\end{aligned}
\end{equation}


\noindent where the input is represented by dynamic motions including a static pose $\mathbf{P_t}$ and its physical dynamics $\mathbf{D_t}$ at time $t$, and a motion encoder $\mathcal{E_M}$ is employed to encode the inputs into motion-dependent features (Sec. \ref{sec:method_me}). Besides, in the training stage, a physical motion decoder $\mathcal{D_M}$ is introduced to enforce the learning of spatial and temporal relations by decoding the intermediate feature $\mathbf{f_t}$ to predict the spatial derivatives at $t+1$ physically corresponding to surface normal $\mathbf{N_{t+1}} =  \frac{\partial \mathbf{P_{t+1}}}{\partial \mathbf{x}}$, and temporal derivatives at $t+1$ corresponding to surface velocity $\mathbf{V_{t+1}} = \frac{\partial \mathbf{P_{t+1}}}{\partial \mathbf{t}}$ (Sec. \ref{sec:method_de}).  $\mathbf{f_t}$ is rendered into human images by a decoder $\mathcal{D_R}$: $\mathcal{G}_2 \circ \mathcal{G}_1$ (Sec. \ref{sec:method_re}). The framework is depicted in Fig. \ref{fig:framework}.

\noindent \textbf{Notation.} In the following parts, we use $\mathbf{f^{space}_t}$ to denote the features in the pipeline, \ie, $\mathbf{f^{3D}_t}$ is a 4D representation that is defined in 3D space with temporal $t$ conditioned, and $\mathbf{M^{3D}_t}$ is 4D motion input defined in 3D space with $t$ conditioned.

\subsection{Surface-based 4D Motion Encoding} \label{sec:method_me}

\noindent \textbf{Extracting 4D Motions.} We first extract 4D motions from a sequence of time-varying parametric posed body (\eg, SMPL) meshes obtained from training video sequences. We describe the motion $\mathbf{M^{3D}_t}$ at time $t$ as a static skeleton pose $\mathbf{P_t}$ and its dynamics $\mathbf{D_t}$. The dynamics at $t$ are physically determined by the current pose, 3D velocity, and motion trajectory of the past several timesteps, which also contribute to the time-varying appearances of the secondary motion. The pose $\mathbf{P_t}$ is represented by the 3D vertices of the posed mesh, and dynamics $\mathbf{D_t}$ are parameterized by 1) body surface velocity $\mathbf{V_t}$ corresponding to the temporal derivatives of the current pose at $t$, and 2) motion trajectory $\mathbf{T_t}$ that aggregates the temporal derivatives over the past several timesteps with a sliding window size of $w$ with weight $\lambda$:
\begin{equation}
\label{eq:form_dy}
\begin{aligned}
    \mathbf{M^{3D}_t} &= \left[ \mathbf{P_t}, \mathbf{D_t} \right], &\mathbf{D_t} & = \left[ \mathbf{V_t}, {\mathbf{T_t}} \right] \\ 
    \mathbf{V_t} & = \frac{\partial \mathbf{P_t}}{\partial \mathbf{t}}, &\mathbf{T_{t}} & = \mathbf{P_t} + \frac{1}{\sum\lambda_i}\sum_{i=1}^w \mathbf{V_{t-i}} * \lambda_i
\end{aligned}
\end{equation}

Note the motion trajectory aggregated from several consecutive timesteps makes the motion representation robust to pose estimation errors, \ie, the pose estimations for two consecutive timesteps may be the same due to pose estimation errors in practice. An ablation study of the motion trajectory representation can be found in the supp. mat.

\noindent \textbf{Recording 4D Motions on the Surface Manifold.} Modeling the motions with temporal dynamics often requires dense observations to construct a 4D motion volume. Instead, we notice that non-rigid deformations of human geometry in motion typically occur around the body surface instead of a 3D volume, and hence we propose to model the motions on the human body surface. Specifically, we project the 4D motion input $\mathbf{M^{3D}_t}$ including spatial pose and temporal dynamics from 3D space into a compact spatially aligned UV space using the geometric transformation $\mathcal{W}$ that is pre-defined by the parametric (\eg, SMPL) body template, which yields $\mathbf{f^{uv}_t} = \mathcal{W} \mathbf{M^{3D}_t}$. The UV-aligned motion feature $\mathbf{f^{uv}_t}$ faithfully preserves the articulated structures of body topology in a compact 2D space.

\noindent \textbf{Generating Surface-based Triplanes for Motion Modeling.} To model clothed humans, we further employ a motion encoder $\mathcal{E_M}$ that lifts the clothless 2D features $\mathbf{f^{uv}_t}$ to a 3D triplane representation $\mathbf{f^{uvh}_t}$ that is defined in a $\mathbf{u-v-h}$ system, with $\mathbf{u-v}$ to represent the motion of the clothless body template, and an extensional coordinate $\mathbf{h}$ to represent the secondary motion of clothes, \ie, the temporal cloth offsets, and hence 4D clothed motions can be parameterized by a surface-based triplane by:
\begin{equation}
\label{eq:fuvh}
    \mathbf{f^{uvh}_t} = \mathcal{E_M}(\mathbf{f^{uv}_t}) = \mathcal{E_M}(\mathcal{W} \mathbf{M^{3D}_t}),
\end{equation}
\noindent where $\mathcal{W}$ is the geometric transformation from 3D space to UV space.
$\mathbf{f^{uvh}_t}$ consists of three planes $\bm{x}_{uv}, \bm{x}_{uh}, \bm{x}_{hv} \in \mathbb{R}^{U\times V\times C}$ to form the spatial relationship, where $U, V$ denote spatial resolution and $C$ is the channel number. In contrast to the volumetric triplanes used in EG3D \cite{Chan2021EfficientG3} where the three planes represent three vertical planes in the 3D volumetric space, $\mathbf{f^{uvh}_t}$ is defined on the human body surface inheriting human topology priors for motion modeling, as illustrated by an unwarped triplane in 3D space in Fig \ref{fig:framework}. 

\subsection{Physical Motion Decoding} \label{sec:method_de}

\noindent We propose a physical motion decoder $\mathcal{D_M}$ to learn spatial and temporal features of motion, which is achieved by decoding the intermediate motion feature $\mathbf{f^{uvh}_t}$ learned by $\mathcal{E_M}$ to predict the motions at the next timestep $t+1$, such as spatial derivatives corresponding to surface normal $\mathbf{N^{uv}_{t+1}}$, temporal derivatives corresponding to surface velocity $\mathbf{V^{uv}_{t+1}}$. Note that we model the normal and velocity in the UV space by: 
\begin{equation}
    \{\mathbf{N^{uv}_{t+1}}, \mathbf{V^{uv}_{t+1}} \}= \mathcal{D_M}(\mathbf{f^{uvh}_t}) = \mathcal{D_M} \circ \mathcal{E_M}(\mathcal{W} \mathbf{M^{3D}_t})
\end{equation}

\figCmpMotionMain

\subsection{4D Appearance Decoding} \label{sec:method_re}
\noindent \textbf{Volumetric Surface-conditioned Rendering.} Given a target camera viewpoint, the triplane $\mathbf{f^{uvh}_t}$ is first rendered into low-resolution volumetric features $\mathbf{I_F} \in \mathbb{R}^{H\times W \times \hat{C}}$ by a conditional NeRF $\mathbb{F}_{\Phi}$, where ${H}$ and $W$ are the spatial resolution and $\hat{C}$ is the channel number. To be more specific, given a 3D query point $\bm{p}_i$, we transform it into a surface-based local coordinate $\hat{\bm{p}_i}=(\bm{u}_i,\bm{v}_i,\bm{h}_i)$ w.r.t. the tracked body mesh $\mathbf{P_t}$. Here we search the nearest face $\bm{f}_i$ of $\mathbf{P_t}$ for query point $\bm{p}_i$ and $(\bm{u}_i, \bm{v}_i)$ and $\bm{h}_i$ are the barycentric coordinates of the nearest point on $\bm{f}_i$, and the signed distance respectively. We therefore obtain the local feature $\bm{z}^{uvh}_i$ of the query point $\bm{p}_i$ as
\begin{equation}
\begin{aligned}    
        \bm{z}^{uvh}_i = \text{cat}\left[\Pi(\bm{x}_{uv}; \bm{u}_i, \bm{v}_i), \Pi(\bm{x}_{uh}; \bm{u}_i, \bm{h}_i), \Pi(\bm{x}_{hv}; \bm{h}_i, \bm{v}_i)\right],
    \end{aligned}
\end{equation}
where $\Pi(\cdot)$ denotes sampling operation, $\text{cat}[\cdot]$ denotes the concatenation operator.


Given camera direction $\bm{d}_i$, the appearance features $\bm{c}_i$ and density features $\bm{\sigma}_i$ of point $\bm{p}_i$ are predicted by
\begin{equation} \label{eq:nerf}
    \{\bm{c}_i, \bm{\sigma}_i\} = \mathbb{F}_{\Phi}(\bm{d}_i, \bm{z}^{uvh}_i)
\end{equation}

We integrate all the radiance features of sampled points into a 2D feature map $\bm{I_F}$ at time $t$ through volume renderer $\mathcal{G}_1$ \cite{Kajiya1984RayTV}
\begin{equation} \label{eq:reslow}
\begin{aligned}
    \mathbf{I_F} = \mathcal{G}_1(\mathbf{f^{uvh}_t}, cam; \mathbb{F}_{\Phi})
\end{aligned}
\end{equation}
\noindent where $\mathbf{f^{uvh}_t}$ is extracted from Eq. \ref{eq:fuvh}.

\noindent \textbf{Efficient Geometry-Aware Super-Resolution}.
Sampling dense points to render the full-resolution volumetric features is computationally heavy. Instead, we employ a super-resolution network $\mathcal{G}_2$~\cite{Chan2021EfficientG3} to render high-resolution images $\mathbf{I^{+}_{RGB}} \in \mathbb{R}^{\hat{H}\times\hat{W}\times 3}$: 

\begin{equation} \label{eq:reslow}
    \mathbf{I^{+}_{RGB}}=\mathcal{G}_2 \circ \mathcal{G}_1(\mathbf{f^{uvh}_t}, cam; \mathbb{F}_{\Phi})
\end{equation}

\subsection{Optimization} \label{sec:method_opt}

\noindent \nickname{} is trained end-to-end to optimize $\mathcal{E_M}$, $\mathcal{D_M}$, and renderers $\mathcal{G}_1$, $\mathcal{G}_1$ with 2D image loss. We employ Adversarial Loss and Reconstruction Loss including Pixel Loss, Perceptual Loss, Face Identity Loss, Velocity and Normal Loss, and Volume Rendering Loss for supervision in training, with Adam \cite{adam} as the optimizer. Refer to the supp. mat. for more details.

\figCmpMotionZ

\section{Experiments}  
\label{sec:exp}

\noindent \textbf{Dataset and Metrics}. We evaluate the novel view synthesis on ZJU-MoCap, MPII-RDDC, and AIST++, with a total of 9 subjects. We use the same camera setup as Neural Body, that is, 4 cameras used for training, and the other 18 for testing. For MPII-RDDC, 18 cameras in training, and 9 for testing. Since ASIT++ only has 9 cameras, we use 6 for training, and the remaining 3 for testing. Refer to more details in the supp. mat. We compare each method on per-pixel metrics including SSIM \cite{ssim} and PSNR, and perception metrics including LPIPS \cite{lpip} and FID \cite{Heusel2017GANsTB}.

\noindent \textbf{Baselines}. We compare our method against SOTA methods including Neural Body \cite{neuralbody}, HumanNeRF \cite{Weng2022HumanNeRFFR}, Instant-NVR \cite{instant_nvr}, ARAH \cite{ARAH:2022:ECCV}, DVA \cite{Remelli2022DrivableVA} and HVTR++ \cite{hvtrpp}. Neural Body models human poses in 3D space with point cloud as pose representation, HumanNeRF and Instant-NVR model poses in a canonical space with inverse skinning, ARAH is a forward-skinning based method, DVA and HVTR++ encode poses in UV space, and take driving view signals as input for telepresence applications. 

\subsection{Comparisons to SOTA Methods}

\tabCmpZju
\noindent \textbf{Quantitative comparisons} We conduct the quantitative comparisons on the three datasets with a total of 9 subjects. The quantitative results on ZJU-MoCap are summarized in Tab. \ref{tab:cmp_zju_sum}, which suggests that our new paradigm outperforms the SOTA by a big margin, and we achieve the best quantitative results on all four metrics. The detailed comparisons on each subject of ZJU-MoCap, and quantitative results on MPII-RDDC and AIST++ are listed in the supp. mat. Refer to the comparisons with ARAH \cite{ARAH:2022:ECCV}, DVA \cite{Remelli2022DrivableVA} and HVTR++ \cite{hvtrpp} in the supp. mat.

\noindent \textbf{Time-varying Appearances with Dynamics Conditioning on ZJU-MoCap.} We compare with baseline methods for novel view synthesis on two motion sequences of subject S313 from the ZJU-MoCap dataset, as shown in Fig. \ref{fig:cmp_313}. We evaluate the capability of each method to synthesize time-varying appearances. Specifically, two sequences S1( swing arms left to right) and S2 (raise and lower arms) are evaluated, where similar poses occur at different timesteps with different dynamics (\eg, movement directions, trajectory, or velocity) are marked in the same color, such as the pairs of \cnum{1}\cnum{2}, \cnum{3}\cnum{4}, and \cnum{5}\cnum{6}. Fig. \ref{fig:cmp_313} suggests that Instant-NVR \cite{instant_nvr} fails to synthesize time-varying high-frequency wrinkles, and Neural Body \cite{neuralbody} synthesizes time-varying yet blurry wrinkles. HumanNeRF \cite{Weng2022HumanNeRFFR} synthesizes detailed yet static T-shirt wrinkles, \ie, the wrinkles are almost the same for similar poses, such as \cnum{1}\cnum{2}, \cnum{3}\cnum{4}, and \cnum{5}\cnum{6}. In contrast, our method renders both high-fidelity and time-varying wrinkles. The comparisons on other subjects S387 and S315 are shown in Fig. \ref{fig:cmp_zju_2}, which illustrate the effectiveness of our proposed paradigm in dynamics learning. 

\figMpiLighting
\noindent \textbf{Time-varying Appearances with Motion-dependent Shadows on MPII-RDDC.} 
We compare with baseline methods for novel view synthesis on MPII-RDDC, as shown in Fig. \ref{fig:lighting}. The sequence is captured in a  studio with top-down lighting that casts shadows on the human body due to self-occlusions. We notice that the synthesis of shadow can be formulated as learning motion-dependent shadows in the motion representation without the need to explicitly model the lighting.  We validate our method from two aspects of the scenario in Fig. \ref{fig:lighting}. \textbf{1)} In Row 1, Fig. \ref{fig:lighting} indicates that with the expressive motion representation and physical motion learning, \nickname{} succeeds in predicting the motion-dependent shadows, such as \cnum{1}\cnum{2}\cnum{3}\cnum{4}, whereas SOTA HumanNeRF renders almost the same T-shirt appearance. \textbf{2)} In Row 2, we compare the synthesis of time-varying appearances for similar poses occurring in a jump-up-and-down motion sequence \ie, \cnum{5} vs. \cnum{6}, which shows that \nickname{} is capable of predicting the clothes offsets under similar pose with different motion trajectories. In addition, we synthesize the shadows \cnum{7}\cnum{8} in the challenging jump-up-and-down motion sequence. However, HumanNeRF fails to predict the dynamic clothing offsets \cnum{9}.

\figAistb
\noindent \textbf{Time-varying Appearances for Fast Motions on AIST++.} We also evaluate our methods by rendering humans with fast dance motions (S21 and S13) on AIST++, as shown in Fig. \ref{fig:aist}, where we compare ours against Neural Body in synthesizing time-varying wrinkles of the T-shirt for different motions. Fig. \ref{fig:aist} suggests that Neural Body can only synthesize blurry results partly because Neural Body models pose in a sparse 3D space. Instead, we model motions in a denser body surface space, which enables high-fidelity image synthesis. Note that AIST++ is based on SMPL tracking with a scaling factor that affects the inverse LBS used in Instant-NVR or HumanNeRF both relying on inverse LBS, and hence we cannot compare them based on the released official code. Quantitative results can be found in the supp. mat.

\figAbTrip

\figMotionCond
\tabAbMotion

\subsection{Ablation Study}

\noindent \textbf{Surface-based Triplane vs. Volumetric Triplane.} We compare the well-adopted volumetric triplane (Vol-Trip) \cite{Chan2021EfficientG3} and our proposed surface-based triplane (Surf-Trip) for human modeling on two subjects S313 and S387 from ZJU-MoCap, as shown in Fig. \ref{fig:ab_trip}. Fig. \ref{fig:ab_trip} (left) suggests that the Surf-Trip converges faster in training with a smaller reconstruction loss for both sequences, and the performances are further improved with physical motion learning, \eg, `Surf-Trip + ML'. Fig. \ref{fig:ab_trip} (right) compares the rendering results qualitatively, which indicates that Surf-Trip is more effective in handling self-occlusions, whereas Vol-Trip fails \cnum{1}\cnum{2}\cnum{3}\cnum{4}, though it performs well in another viewpoint without self-occlusions \cnum{6}. In addition, Surf-Trip generates high-fidelity clothing wrinkles and facial details, whereas the face rendering of Vol-Trip is blurry.

\noindent \textbf{Motion Conditioning and Learning.} We also analyze the effectiveness of our proposed motion conditioning and learning qualitatively and quantitatively. Tab. \ref{tab:AbMotion} summarizes the quantitative results, which suggest that the quantitative results are significantly improved by taking as input the dynamics conditioning $D_{cond}$. With physical motion learning, \ie, predicting the temporal derivatives surface velocity and spatial derivates normal at the next timestep, the quantitative results are further improved by a big margin. 

The qualitative comparisons are shown in Fig. \ref{fig:ab_motion}, where in Row 1, we observe higher-fidelity appearances with dynamics conditioning, and physical motion learning for both velocity and normal prediction. In Row 2, we evaluate whether our method learns to decouple the temporal dynamics (\eg, velocity) and static poses from motion conditioning, which corresponds to the rendering of clothing wrinkles of secondary motion (T-shirt \cnum{1}) and tight parts (\eg, shorts \cnum{2}). We only change the velocity for each variant in the ablation study. It suggests that the appearance of the T-shirt varies when the velocity or dynamics are changed, whereas the renderings of tight parts remain the same, such as the head, tight shorts, and shoes. This is consistent with our daily observations. The ablation study illustrates that our method decouples the pose and dynamics, and is capable of generating dynamics-dependent appearances. Note that the appearance changes are small between $V*1$ and $V*0.01$ when we reduce the velocity, partly because the method is not sensitive to smaller velocity due to the pose estimation errors in the training dataset where consecutive frames may be estimated with the same pose fitting.



\section{Discussion}


\noindent We propose \nickname{}, a new paradigm for learning dynamic humans from videos by jointly modeling temporal motions and human appearances in a unified framework, based on an efficient surface-based triplane. We conduct a systematical analysis of how human appearances are affected by temporal dynamics, and extensive experiments validate the expressiveness of the surface-based triplane in rendering fast motions and motion-dependent shadows. Quantitative experiments illustrate that \nickname{} achieves the SOTA results on different motion sequences.



\section*{Acknowledgement}

This study is supported by the Ministry of Education, Singapore, under its MOE AcRF Tier 2 (MOET2EP20221- 0012), NTU NAP, and under the RIE2020 Industry Alignment Fund – Industry Collaboration Projects (IAF-ICP) Funding Initiative, as well as cash and in-kind contribution from the industry partner(s).

\newcommand{\figSupTrip}{
\begin{figure*}[th]
	\begin{center}
		\includegraphics[width=\linewidth]{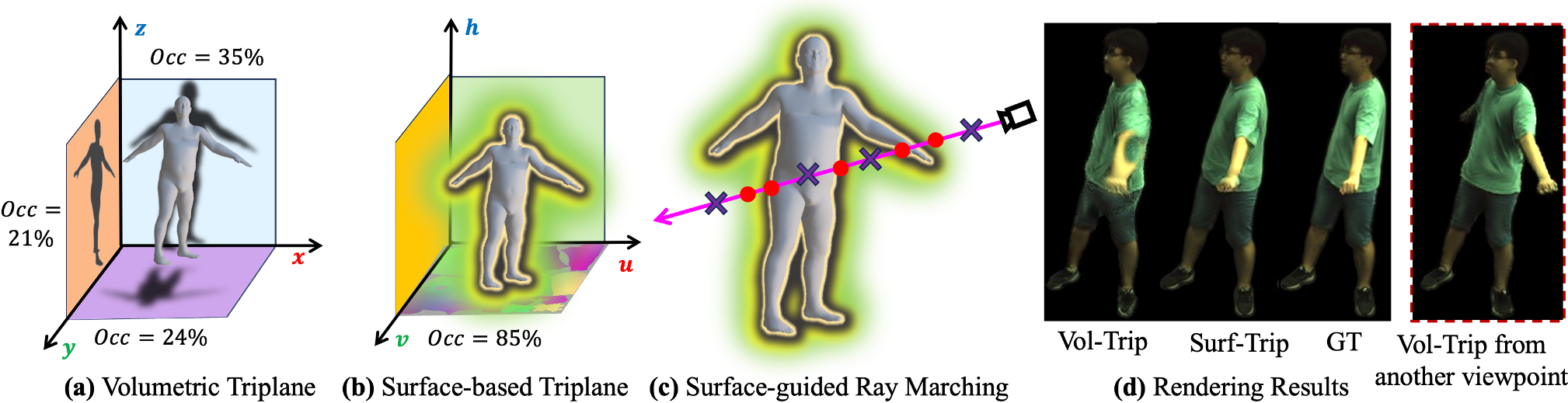}
	\end{center}
	\caption{Illustration of Volumetric Triplane vs. Surface-based Triplane.}
	\label{fig:sup_trip}

\end{figure*}
}

\newcommand{\figSupCmpDVA}{
\begin{figure*}[th]
	\begin{center}
		\includegraphics[width=\linewidth]{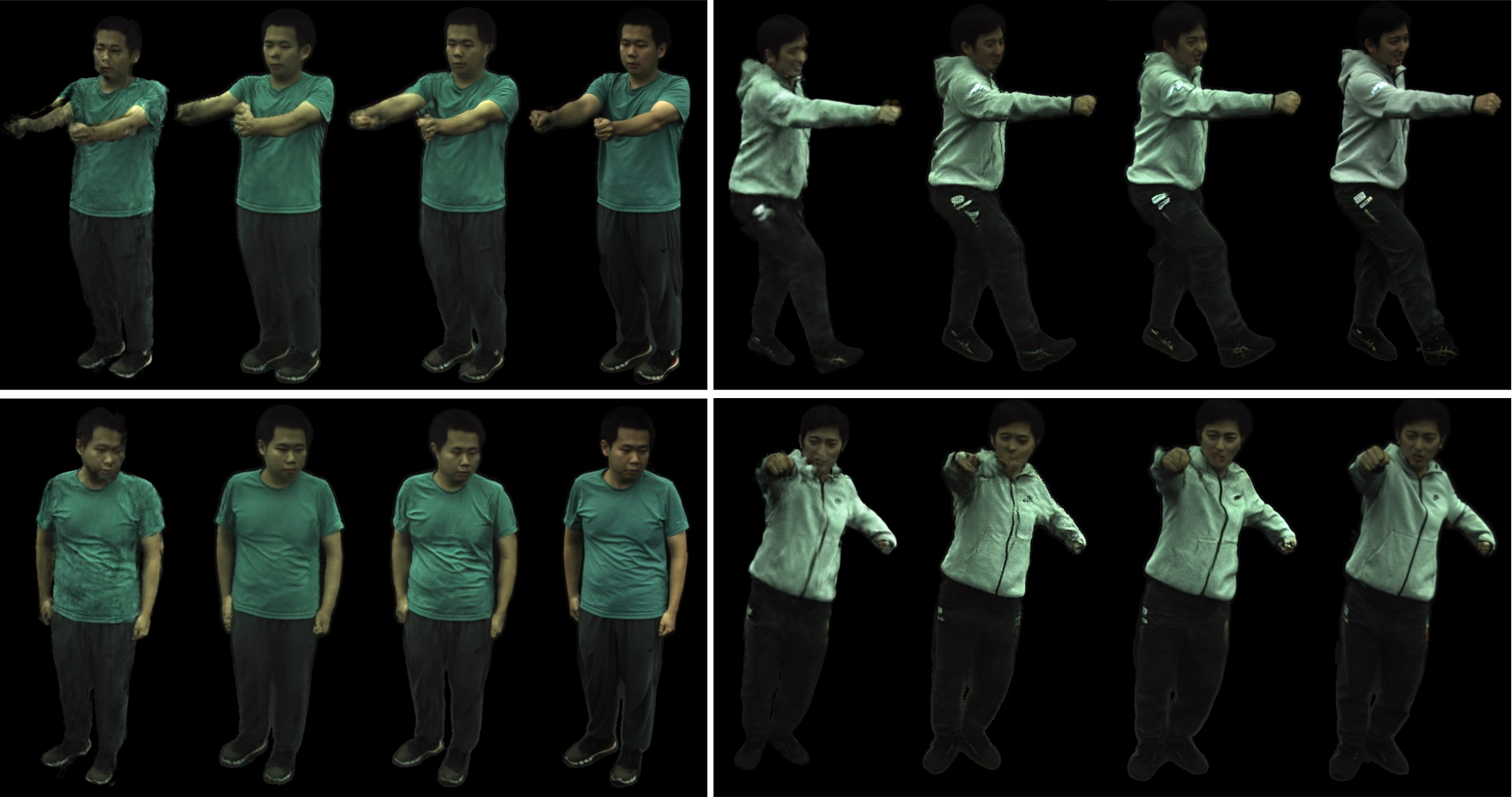}
	\end{center}
	\caption{Qualitative comparisons against the 3D pose- and image-driven approach DVA \cite{Remelli2022DrivableVA} and HVTR++ \cite{hvtrpp} for novel view synthesis of training poses on ZJU-MoCap. For each example, from left to right: DVA, HVTR++, Ours, Ground Truth. Rendering results of DVA and HVTR++ are provided by the authors.}
	\label{fig:sup_cmpdva}
\end{figure*}
}

\newcommand{\figSupCmpPose}{
\begin{figure}[]
	\begin{center}
		\includegraphics[width=\linewidth]{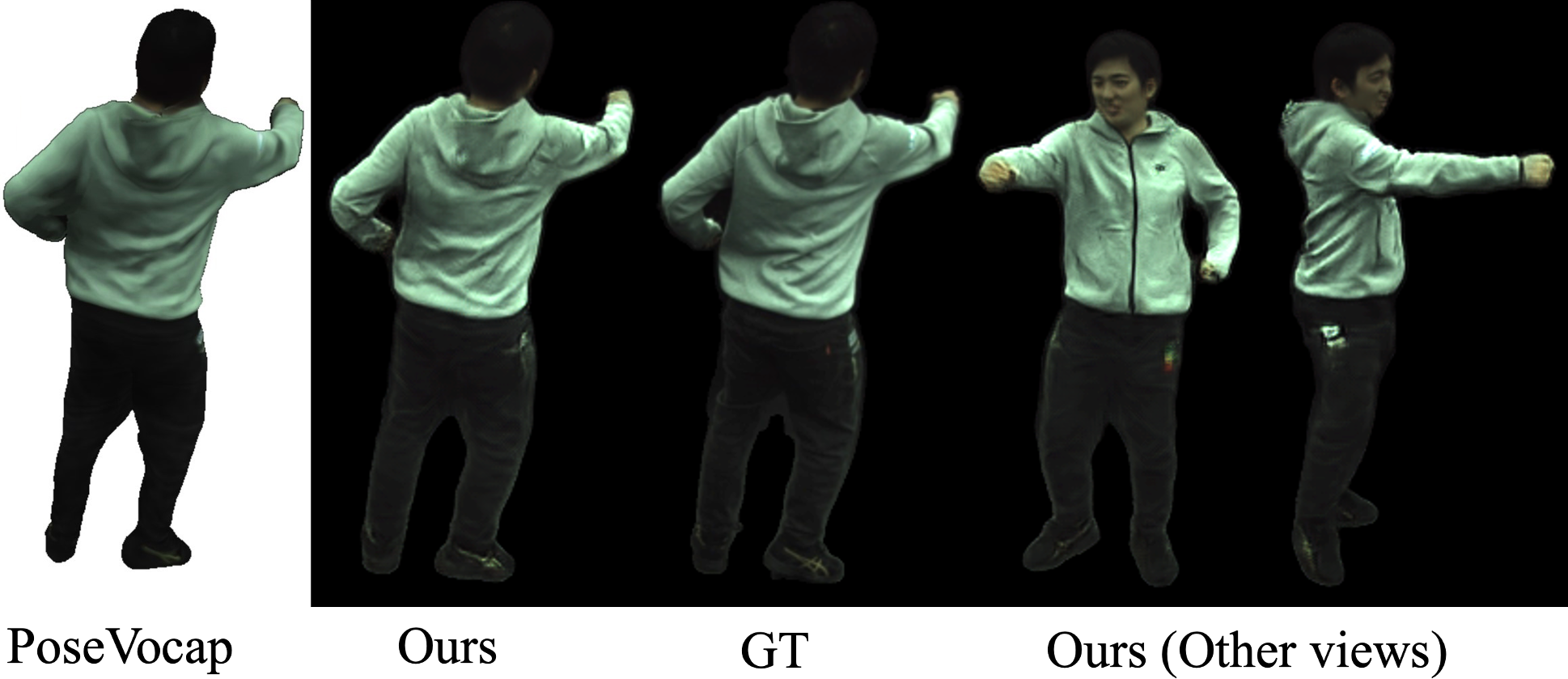}
	\end{center}
	\caption{Qualitative comparisons against PoseVocap \cite{li2023posevocab}  for novel view synthesis of training poses on ZJU-MoCap.}
	\label{fig:sup_cmpposevocap}
\end{figure}
}

\newcommand{\figSupCmpArah}{
\begin{figure}[]
	\begin{center}
		\includegraphics[width=\linewidth]{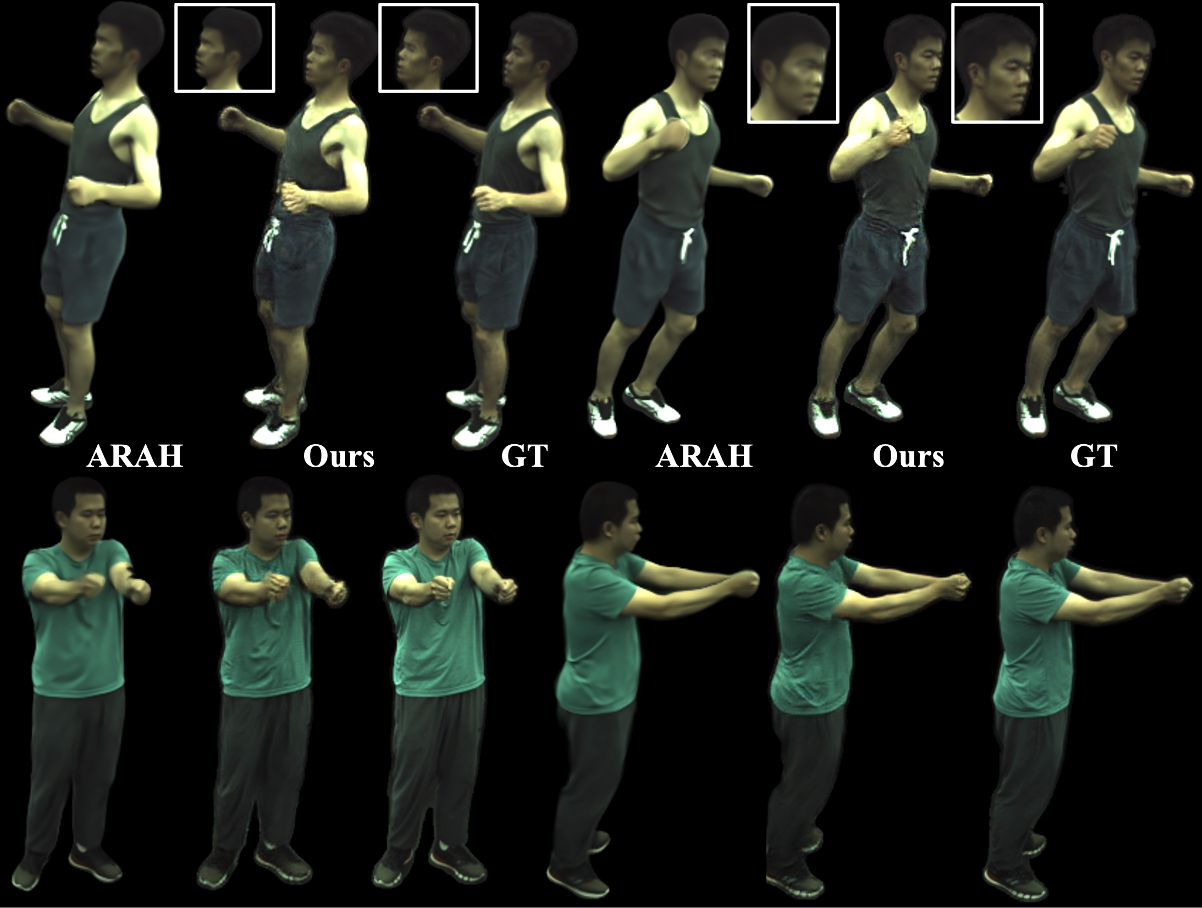}
	\end{center}
	\caption{Qualitative comparisons against ARAH \cite{ARAH:2022:ECCV} for novel view synthesis of novel poses on ZJU-MoCap.}
	\label{fig:sup_cmparah}
\end{figure}
}

\newcommand{\figSupFail}{
\begin{figure}[th]
	\begin{center}
		\includegraphics[width=\linewidth]{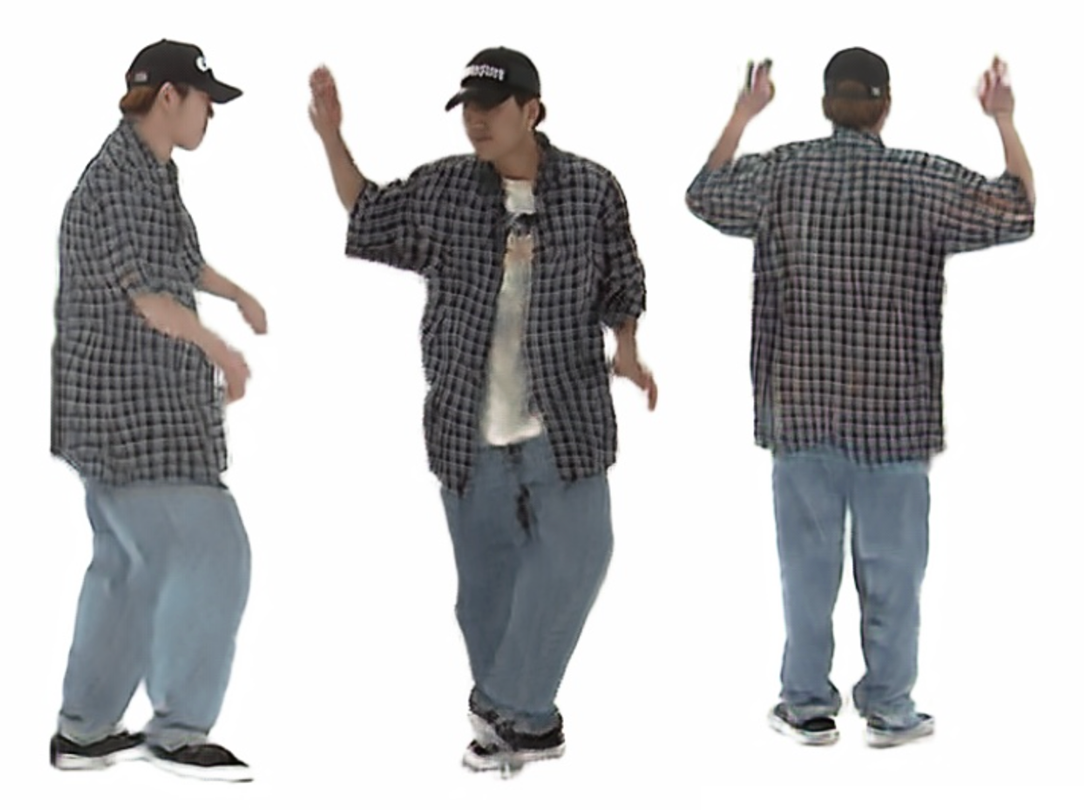}
	\end{center}
	\caption{Failure cases.}
	\label{fig:sup_fail}
\end{figure}
}

\clearpage
\newpage

\noindent \appendixdef

\appendix
\renewcommand{\thesection}{\Alph{section}}

\figSupTrip
\section{Implementation} 

\subsection{Network Architectures}

\noindent \textbf{Motion Encoder and Decoder.}
The motion encoder is based on the Pix2PixHD \cite{pix2pixhd} architecture with 3 Encoder blocks of [Conv2d, Batch- Norm, ReLU], ResNet \cite{He2016DeepRL} blocks, and 3 Decoder blocks of [ReLU, ConvTranspose2d, BatchNorm]. The motion decoder has 2 Decoder blocks.

\noindent \textbf{Volume Renderer.} We use a 5-layer MLP with a skip connection from the input to the 3th layer as in DeepSDF \cite{park2019deepsdf}. From the 4th layer, the network branches out two heads, one to predict density with one fully-connected layer and the other one to predict color features with two fully-connected layers.

\noindent \textbf{Super-Resolution.} To super-resolve low-resolution volumetric features to low-resolution images, we first bilinearly upsample the features by 2$\times$ and then feed the upsampled features into two convolutional layers with a kernel size of 3 to upsample the images by a factor of 2.

\noindent \textbf{Surface-based Triplane.} The size of the triplane is $256\times256\times48$.  
 
\noindent \textbf{Discriminator.} We adopt the discriminator architecture of PatchGAN \cite{pix2pix} for adversarial training. Note that different from EG3D \cite{eg3d} that applies the image discriminator at both resolutions, we only supervise the final rendered images with adversarial training and supervise the volumetric features with reconstruction loss.

\subsection{Optimization} \label{sec:method_opt_sup}

\nickname{} is trained end-to-end to optimize $\mathcal{E_M}$, $\mathcal{D_M}$, and renderers $\mathcal{G}_1$, $\mathcal{G}_1$ with 2D image loss. Given a ground truth image $I_{gt}$, we predict a target RGB image $\mathbf{I^{+}_{RGB}}$ with the following loss: 

\noindent \textbf{Pixel Loss}. We enforce an $\ell_1$ loss between the generated image and ground truth as ${L}_{pix} = \|I_{gt} - \mathbf{I^{+}_{RGB}} \|_1$.

\noindent \textbf{Perceptual Loss}. Pixel loss is sensitive to image misalignment due to pose estimation errors, and we further use a perceptual loss \cite{Perceptual_Losses} to measure the differences between the activations on different layers of the pre-trained VGG network \cite{vgg} of the generated image $\mathbf{I^{+}_{RGB}}$ and ground truth image $I_{gt}$,
\begin{equation}
	{L}_{vgg}=\sum \frac{1}{N^{j}}\left\|g^{j}\left(I_{gt}\right)-g^{j}\left(\mathbf{I^{+}_{RGB}}\right)\right\|_2,
\end{equation}
\noindent where $g^j$ is the activation and $N^j$ the number of elements of the $j$-th layer in the pretrained VGG network.

\noindent \textbf{Adversarial Loss}. We leverage a multi-scale discriminator $D$ \cite{pix2pixhd} as an adversarial loss ${L}_{adv}$ to enforce the realism of rendering, especially for the cases where estimated human poses are not well aligned with the ground truth images.

\noindent \textbf{Face Identity Loss}. We use a pre-trained network to ensure that the renderers preserve the face identity on the cropped face of the generated and ground truth image,
\begin{equation}
	{L}_{{face}}= \| N_{ {face}}\left(I_{gt}\right)-N_{ {face}}\left(\mathbf{I^{+}_{RGB}}\right)\|_2,
\end{equation}
\noindent where $N_{face}$ is the pretrained SphereFaceNet \cite{Liu2017SphereFaceDH}.

\noindent \textbf{Velocity Loss}. We employ a velocity loss (temporal motion derivates) for the motion dececoding supervision, 
\begin{equation}
	{L}_{{velocity}}= \| V^{uv}_{gt(t+1)} - \mathbf{V^{uv}_{t+1}} \|_2,
\end{equation}
\noindent where $V^{uv}_{gt(t+1)}$ is the ground truth velocity at timestep $t+1$, and $\mathbf{V^{uv}_{t+1}}$ is the predicted velocity by $\mathcal{D_M}$ at timestep $t+1$.

\noindent \textbf{Normal Loss}. We also employ a surface normal loss (spatial motion derivates) for the motion dececoding supervision,
\begin{equation}
	{L}_{{normal}}= \| N^{uv}_{gt(t)} - \mathbf{N^{uv}_{t}} \|_2,	
\end{equation}
\noindent where $N^{uv}_{gt(t)}$ is the ground truth normal at timestep $t$, and $\mathbf{N^{uv}_{t}}$ is the predicted normal by $\mathcal{D_M}$ at timestep $t$. Note that in practical implementation, $\mathcal{D_M}$ first predicts $\mathbf{N^{uv}_{t}}$, which is easier for the network than predicting $\mathbf{N^{uv}_{t+1}}$ directly, and $\mathbf{N^{uv}_{t + 1}}$ can be drived and normalized from:  $\mathbf{N^{uv}_{t + 1}} = \frac{\partial \mathbf{P^{uv}_{t+1}}}{\partial \mathbf{x}} = \frac{\partial \{\mathbf{P^{uv}_{t}} + \mathbf{V^{uv}_{t+1}}\}}{\partial \mathbf{x}} = \mathbf{N^{uv}_{t}} + \frac{\partial \mathbf{V^{uv}_{t+1}}}{\partial \mathbf{x}}$. With the $\mathbf{V^{uv}_{t+1}}$ predicted for temporal motion supervision, the prediction of $\mathbf{N^{uv}_{t}}$ enforces a similar supervision with $\mathbf{N^{uv}_{t+1}}$ for the spatial motion learning.

\noindent \textbf{Volume Rendering Loss}. We supervise the training of volume rendering at low resolution, which is applied on the first three channels of $\mathbf{I_F}$, ${L}_{vol} = \|\mathbf{I_F}[:3] - I^D_{gt}\|_2$. $I^D_{gt}$ is the downsampled reference image. 

The networks were trained using the Adam optimizer \cite{adam}. The loss weights \{$\lambda_{pix}$, $\lambda_{vgg}$, $\lambda_{adv}$, $\lambda_{face}$, $\lambda_{velocity}$, $\lambda_{normal}$, $\lambda_{vol}$ \} are set empirically to $\{.5, 10, 1, 5, 1, 1, 15\}$. It takes about 12 hours to train a model from about 3000 images with 200 epochs on two NVIDIA V100 GPUs.

\subsection{Training Data Processing.} We evaluate the novel view synthesis on three datasets: ZJU-MoCap \cite{neuralbody} (including sequences of S313, S315, S377, S386, S387, S394) at resolution $1024\times 1024$, MPII-RDDC \cite{surreal} at resolution $1285\times940$, and AIST++ \cite{aist} at $1920\times1080$. Note that sequences of ZJU-MoCap used in Neural Body are generally short, \eg, only 60 frames for S313. Instead, to evaluate the time-varying effects, we extend the original training frames of S313, S315, S387, S394 to 400, 700, 600, 600 frames respectively depending on the pose variance of each sequence, whereas S377 and S386 remain the same 300 frames as the setup of Neural Body \cite{neuralbody}. 4 cameras are used for training, and the others are used in testing for ZJU-MoCap. 6 cameras are used in training, 3 for testing in AIST++, 18 cameras for training and 9 cameras for testing in MPI-RDDC.

\figSupCmpDVA

\section{Additional Experimental Results}

\subsection{Comparisons with SOTA Methods}

\tabcmpdva
\noindent \textbf{Comparisons with 3D pose- and image-driven approaches.} In contrast to pose-driven methods (\eg, Neural Body \cite{neuralbody}, Instant-NVR \cite{instant_nvr}, HumanNeRF \cite{Weng2022HumanNeRFFR}), DVA \cite{Remelli2022DrivableVA} and HVTR++ \cite{hvtrpp} propose to utilize both the pose and driving view features in rendering. They model both the pose and texture features in UV space, whereas ours is distinguished by modeling motions in a surface-based triplane, and we jointly learn physical motions and rendering in a unified network for faithful rendering.

Tab. \ref{tab:cmp_dva} summarizes the quantitative results for novel view synthesis on the two sequences (S386 and S387) mentioned in DVA, which suggest that our method significantly outperforms DVA and HVTRPP in terms of both per-pixel and perception metrics. Qualitative comparisons are provided in Fig. \ref{fig:sup_cmpdva}, which shows that our method produces sharper reconstructions with faithful wrinkles than both DVA and HVTR++. In contrast to the image resolution of 512 × 512 used in Neural Body \cite{neuralbody}, HumanNeRF \cite{Weng2022HumanNeRFFR} and Instant-NVR \cite{instant_nvr}, DVA and HVTR++ were trained and evaluated at the resolution of 1024 × 1024 in \cite{Remelli2022DrivableVA,hvtrpp}. We follow the same protocol used in \cite{Remelli2022DrivableVA,hvtrpp} for fair comparisons.

\figSupCmpPose
\noindent \textbf{Comparisons with PoseVocap \cite{li2023posevocab}.}
PoseVocap \cite{li2023posevocab} proposes joint-structured pose embeddings for better temporal consistency in rendering. Qualitative comparisons on novel view synthesis are shown in Fig. \ref{fig:sup_cmpposevocap}, which suggest that our method is capable of generating higher-quality wrinkles than PoseVocap \cite{li2023posevocab}. Note that PoseVocap only provides qualitative results on ZJU-MoCap, and the test results of PoseVocap are reported in the paper \cite{li2023posevocab}.

\figSupCmpArah

\noindent \textbf{Pose Generalization}. Our method is focused on generating free-viewpoint video of dynamic humans, whereas we evaluate the pose generalization capability on ZJU-MoCap and it is observed that our method is not overfitted to the training poses, as suggested in Fig. \ref{fig:sup_cmparah} and  Tab. \ref{tab:cmp_arah}. 

\tabcmparah
Compared to ARAH \cite{ARAH:2022:ECCV} (a forward-skinning-based approach), the state-of-the-art method in pose generalization tasks, we generate better quantitative results in terms of novel view synthesis on training poses or novel poses as summarized in Tab. \ref{tab:cmp_arah}. The qualitative comparisons in Fig. \ref{fig:sup_cmparah} suggest that our method is capable of synthesizing higher-quality faces and cloth wrinkles than ARAH. Note that our method is not targeted at animation, and since the pose variance of ZJU-MoCap is not big enough, the experiments do not illustrate that our method achieves the SOTA results in animation tasks. However, the experimental results suggest that our method is not completely overfitted to the training poses. We use the publicly released test results of ARAH for comparisons.

\tabcmpMPI
\tabcmpAIST
\subsection{Quantitative Comparisons on MPII-RDDC and AIST++ Datasets.}
\noindent The quantitative comparisons on MPII-RDDC \cite{Habermann2021RealtimeDD} are summarized in Tab. \ref{tab:cmp_mpi}, which suggests that our method outperforms HumanNeRF in the lighting-conditioned scenario. The quantitative comparisons on AIST++ \cite{aist} are summarized in Tab. \ref{tab:cmp_aist}, which confirms the effectiveness of our method in rendering fast motions. 

\subsection{Ablation study}
\noindent \textbf{Surface-based Triplane vs. Volumetric Triplane.} We compare the volumetric triplane (Vol-Trip) \cite{Chan2021EfficientG3} and our proposed surface-based triplane (Surf-Trip) for human modeling as shown in Fig. \ref{fig:sup_trip}. It is observed that the volumetric triplane is a sparse representation for human body modeling, \ie, only 21-35\% features are utilized to render the human under the specific pose, and hence the Vol-Trip fails to handle the self-occlusions effectively as shown in Fig. \ref{fig:sup_trip} (d), though Vol-Trip generates plausible results from another viewpoint without sever self-occlusions. In contrast, about 85\% surface-based triplane features are utilized in rendering. In addition, with surface-guided ray marching, our method is more efficient by filtering out invalid points that are far from the body surface.

\tabAbSupMotion
\noindent \textbf{Motion Prediction.} Predicting the next frame based on the status of the current frame is a one-to-many mapping problem. However, we take as input additional dynamics, and trajectory features to infer the motion of the next frame, which alleviates the one-to-many mapping issue. The paper is not focused on motion prediction/generation. Instead, we use {the motion prediction to force a meaningful embedding of the feature space}, which improves the rendering quality. {Predicting the next motion frame $Pred_{t+1}$ offers higher-quality rendering than predicting the current motion frame $Pred_{t}$}, \ie $\mathbf{V^{uv}_{t+1}}$ vs. $\mathbf{V^{uv}_{t}}$, as listed in  Tab. \ref{tab:absupMotion}. We conduct experiments on the S313 and S387 sequences of the ZJU-MoCap dataset in Tab. \ref{tab:absupMotion}.

\noindent \textbf{Training Views.} Tab. \ref{tab:absupMotion} suggests that the performances of novel view synthesis degrade with fewer training views, \ie, from 4 training views $Pred_{t+1}$ to 1 view $Pred_{t+1}(1 ~view)$. Even with 1 view, our performance is still comparable with Instant-NVR (Tab. \ref{tab:cmp_zju_all}).

\tabcmpDynamics
\noindent \textbf{Dynamics Conditioning.} We compare the methods of conditioning dynamics in the rendering network between \cite{Yoon2022LearningMA} and ours. \cite{Yoon2022LearningMA} takes as input the velocities of the past 10 consecutive poses and normal maps of the current pose, whereas we take as input the positional map of the current pose and aggregated trajectory of the past 5 frames as input. Tab. \ref{tab:cmp_dynamics} suggests that our method enables better quantitative results, and we improve the performances by further learning motions, \eg, surface velocity and normal prediction.

\tabcmpSR
\noindent \textbf{Super-resolution.}
Our method utilizes a super-resolution module to synthesize high-quality images. The quantitative results are summarized in Tab. \ref{tab:cmp_sr}. It is observed that the performances are improved when the upsampling factor is increased from 4 to 2, which indicates more geometric features are utilized by increasing the resolution of volumetric rendering.  

\subsection{Efficiency}
\noindent At test time, our method runs at ~3.2 FPS on one NVIDIA V100 GPU to render 512$\times$512 resolution images, about 39$\times$ faster than Neural Body \cite{neuralbody}, 17$\times$ faster than HumanNeRF \cite{Weng2022HumanNeRFFR}, and 9$\times$ faster than Instant-NVR \cite{instant_nvr}.

\figSupFail

\subsection{Failure Cases}
\noindent Our method fails to generate high-quality wrinkles for complicated textures of AIST++ \cite{aist}, as shown in Fig. \ref{fig:sup_fail}. This is because we cannot learn to infer dynamic wrinkles from the complicated appearances.


\tabCmpZjuAll

{\small
\bibliographystyle{ieee_fullname}
\bibliography{main_short}

\begin{thebibliography}{10}\itemsep=-1pt

\bibitem{Aberman2019DeepVP}
Kfir Aberman, M. Shi, Jing Liao, Dani Lischinski, B. Chen, and D. Cohen-Or.
\newblock Deep video‐based performance cloning.
\newblock {\em Computer Graphics Forum}, 38, 2019.

\bibitem{Borshukov2003UniversalCI}
George Borshukov, Dan Piponi, Oystein Larsen, J.~P. Lewis, and Christina Tempelaar-Lietz.
\newblock Universal capture: image-based facial animation for "the matrix reloaded".
\newblock In {\em SIGGRAPH '03}, 2003.

\bibitem{Carranza2003FreeviewpointVO}
Joel Carranza, Christian Theobalt, Marcus~A. Magnor, and Hans-Peter Seidel.
\newblock Free-viewpoint video of human actors.
\newblock {\em ACM SIGGRAPH 2003 Papers}, 2003.

\bibitem{edn}
Caroline Chan, Shiry Ginosar, Tinghui Zhou, and Alexei~A. Efros.
\newblock Everybody dance now.
\newblock {\em ICCV}, pages 5932--5941, 2019.

\bibitem{Chan2021EfficientG3}
Eric Chan, Connor~Z. Lin, Matthew~A. Chan, Koki Nagano, Boxiao Pan, Shalini~De Mello, Orazio Gallo, Leonidas~J. Guibas, Jonathan Tremblay, S. Khamis, Tero Karras, and Gordon Wetzstein.
\newblock Efficient geometry-aware 3d generative adversarial networks.
\newblock {\em ArXiv}, abs/2112.07945, 2021.

\bibitem{eg3d}
Eric~R Chan, Connor~Z Lin, Matthew~A Chan, Koki Nagano, Boxiao Pan, Shalini De~Mello, Orazio Gallo, Leonidas~J Guibas, Jonathan Tremblay, Sameh Khamis, et~al.
\newblock Efficient geometry-aware 3d generative adversarial networks.
\newblock In {\em Proceedings of the IEEE/CVF Conference on Computer Vision and Pattern Recognition}, pages 16123--16133, 2022.

\bibitem{Chen2021AnimatableNR}
Jianchuan Chen, Ying Zhang, Di Kang, Xuefei Zhe, Linchao Bao, and Huchuan Lu.
\newblock Animatable neural radiance fields from monocular rgb video.
\newblock {\em ArXiv}, abs/2106.13629, 2021.

\bibitem{Chen2019LearningIF}
Zhiqin Chen and Hao Zhang.
\newblock Learning implicit fields for generative shape modeling.
\newblock {\em CVPR}, 2019.

\bibitem{Jeruzalski2020NASANA}
Boyang Deng, John~P Lewis, Timothy Jeruzalski, Gerard Pons-Moll, Geoffrey Hinton, Mohammad Norouzi, and Andrea Tagliasacchi.
\newblock Nasa neural articulated shape approximation.
\newblock In {\em ECCV}, 2020.

\bibitem{instant_nvr}
Chen Geng, Sida Peng, Zhen Xu, Hujun Bao, and Xiaowei Zhou.
\newblock Learning neural volumetric representations of dynamic humans in minutes.
\newblock In {\em CVPR}, 2023.

\bibitem{gans}
Ian~J. Goodfellow, Jean Pouget-Abadie, Mehdi Mirza, Bing Xu, David Warde-Farley, Sherjil Ozair, Aaron~C. Courville, and Yoshua Bengio.
\newblock Generative adversarial nets.
\newblock In {\em NIPS}, 2014.

\bibitem{Grigorev2019CoordinateBasedTI}
A.~K. Grigor'ev, Artem Sevastopolsky, Alexander Vakhitov, and Victor~S. Lempitsky.
\newblock Coordinate-based texture inpainting for pose-guided human image generation.
\newblock {\em CVPR}, pages 12127--12136, 2019.

\bibitem{Gu2021StyleNeRFAS}
Jiatao Gu, Lingjie Liu, Peng Wang, and Christian Theobalt.
\newblock Stylenerf: A style-based 3d-aware generator for high-resolution image synthesis.
\newblock {\em ArXiv}, abs/2110.08985, 2021.

\bibitem{Habermann2021RealtimeDD}
Marc Habermann, Lingjie Liu, Weipeng Xu, Michael Zollhoefer, Gerard Pons-Moll, and Christian Theobalt.
\newblock Real-time deep dynamic characters.
\newblock {\em ACM Transactions on Graphics (TOG)}, 40:1 -- 16, 2021.

\bibitem{He2016DeepRL}
Kaiming He, X. Zhang, Shaoqing Ren, and Jian Sun.
\newblock Deep residual learning for image recognition.
\newblock {\em CVPR}, pages 770--778, 2016.

\bibitem{Heusel2017GANsTB}
Martin Heusel, Hubert Ramsauer, Thomas Unterthiner, Bernhard Nessler, and Sepp Hochreiter.
\newblock Gans trained by a two time-scale update rule converge to a local nash equilibrium.
\newblock In {\em NIPS}, 2017.

\bibitem{Hong2021HeadNeRFAR}
Yang Hong, Bo Peng, Haiyao Xiao, Ligang Liu, and Juyong Zhang.
\newblock Headnerf: A real-time nerf-based parametric head model.
\newblock {\em ArXiv}, abs/2112.05637, 2021.

\bibitem{densepointclouds}
Tao Hu, Geng Lin, Zhizhong Han, and Matthias Zwicker.
\newblock Learning to generate dense point clouds with textures on multiple categories.
\newblock In {\em Proceedings of the IEEE/CVF Winter Conference on Applications of Computer Vision (WACV)}, pages 2170--2179, January 2021.

\bibitem{egorend}
Tao Hu, Kripasindhu Sarkar, Lingjie Liu, Matthias Zwicker, and Christian Theobalt.
\newblock Egorenderer: Rendering human avatars from egocentric camera images.
\newblock In {\em ICCV}, 2021.

\bibitem{hvtrpp}
Tao Hu, Hongyi Xu, Linjie Luo, Tao Yu, Zerong Zheng, He Zhang, Yebin Liu, and Matthias Zwicker.
\newblock Hvtr++: Image and pose driven human avatars using hybrid volumetric-textural rendering.
\newblock {\em IEEE Transactions on Visualization and Computer Graphics}, pages 1--15, 2023.

\bibitem{hvtr}
T. Hu, Tao Yu, Zerong Zheng, He Zhang, Yebin Liu, and Matthias Zwicker.
\newblock Hvtr: Hybrid volumetric-textural rendering for human avatars.
\newblock {\em 3DV}, 2022.

\bibitem{Huang2020ARCHAR}
Zeng Huang, Yuanlu Xu, Christoph Lassner, Hao Li, and Tony Tung.
\newblock Arch: Animatable reconstruction of clothed humans.
\newblock {\em 2020 (CVPR)}, pages 3090--3099, 2020.

\bibitem{pix2pix}
Phillip Isola, Jun-Yan Zhu, Tinghui Zhou, and Alexei~A. Efros.
\newblock Image-to-image translation with conditional adversarial networks.
\newblock {\em CVPR}, pages 5967--5976, 2017.

\bibitem{Perceptual_Losses}
Justin Johnson, Alexandre Alahi, and Li Fei-Fei.
\newblock Perceptual losses for real-time style transfer and super-resolution.
\newblock volume 9906, pages 694--711, 10 2016.

\bibitem{Kajiya1984RayTV}
James~T. Kajiya and Brian~Von Herzen.
\newblock Ray tracing volume densities.
\newblock {\em Proceedings of the 11th annual conference on Computer graphics and interactive techniques}, 1984.

\bibitem{adam}
Diederick~P Kingma and Jimmy Ba.
\newblock Adam: A method for stochastic optimization.
\newblock In {\em ICLR}, 2015.

\bibitem{KratzHPV2017}
Bernhard Kratzwald, Zhiwu Huang, Danda~Pani Paudel, and Luc Van~Gool.
\newblock Towards an understanding of our world by {GANing} videos in the wild.
\newblock arXiv:1711.11453, 2017.

\bibitem{aist}
Ruilong Li, Shan Yang, David~A. Ross, and Angjoo Kanazawa.
\newblock Learn to dance with aist++: Music conditioned 3d dance generation, 2021.

\bibitem{li2023posevocab}
Zhe Li, Zerong Zheng, Yuxiao Liu, Boyao Zhou, and Yebin Liu.
\newblock Posevocab: Learning joint-structured pose embeddings for human avatar modeling.
\newblock In {\em ACM SIGGRAPH Conference Proceedings}, 2023.

\bibitem{neuralactor}
Lingjie Liu, Marc Habermann, Viktor Rudnev, Kripasindhu Sarkar, Jiatao Gu, and Christian Theobalt.
\newblock Neural actor: Neural free-view synthesis of human actors with pose control.
\newblock {\em TOG}, 40, 2021.

\bibitem{liu2020NeuralHumanRendering}
Lingjie Liu, Weipeng Xu, Marc Habermann, Michael Zollhöfer, Florian Bernard, Hyeongwoo Kim, Wenping Wang, and Christian Theobalt.
\newblock Neural human video rendering by learning dynamic textures and rendering-to-video translation.
\newblock {\em IEEE Transactions on Visualization and Computer Graphics}, 05 2020.

\bibitem{Liu2019}
Lingjie Liu, Weipeng Xu, Michael Zollhoefer, Hyeongwoo Kim, Florian Bernard, Marc Habermann, Wenping Wang, and Christian Theobalt.
\newblock Neural rendering and reenactment of human actor videos.
\newblock {\em ACM Transactions on Graphics (TOG)}, 2019.

\bibitem{Liu2017SphereFaceDH}
Weiyang Liu, Y. Wen, Zhiding Yu, Ming Li, B. Raj, and Le Song.
\newblock Sphereface: Deep hypersphere embedding for face recognition.
\newblock {\em CVPR}, pages 6738--6746, 2017.

\bibitem{smpl}
M. Loper, Naureen Mahmood, J. Romero, Gerard Pons-Moll, and Michael~J. Black.
\newblock Smpl: a skinned multi-person linear model.
\newblock {\em ACM Trans. Graph.}, 34:248:1--16, 2015.

\bibitem{MaSJSTV2017}
Liqian Ma, Xu Jia, Qianru Sun, Bernt Schiele, Tinne Tuytelaars, and Luc Van~Gool.
\newblock Pose guided person image generation.
\newblock In {\em NeurIPS}, pages 405--415, 2017.

\bibitem{Ma18}
Liqian Ma, Qianru Sun, Stamatios Georgoulis, Luc van Gool, Bernt Schiele, and Mario Fritz.
\newblock Disentangled person image generation.
\newblock {\em CVPR}, 2018.

\bibitem{scale}
Qianli Ma, Shunsuke Saito, Jinlong Yang, Siyu Tang, and Michael~J. Black.
\newblock Scale: Modeling clothed humans with a surface codec of articulated local elements.
\newblock In {\em CVPR}, 2021.

\bibitem{pop}
Qianli Ma, Jinlong Yang, Siyu Tang, and Michael~J Black.
\newblock The power of points for modeling humans in clothing.
\newblock In {\em ICCV}, 2021.

\bibitem{Mescheder2019OccupancyNL}
Lars~M. Mescheder, Michael Oechsle, Michael Niemeyer, Sebastian Nowozin, and Andreas Geiger.
\newblock Occupancy networks: Learning 3d reconstruction in function space.
\newblock {\em CVPR}, 2019.

\bibitem{Michalkiewicz2019DeepLS}
Mateusz Michalkiewicz, Jhony~Kaesemodel Pontes, Dominic Jack, Mahsa Baktash, and Anders~P. Eriksson.
\newblock Deep level sets: Implicit surface representations for 3d shape inference.
\newblock {\em ArXiv}, 2019.

\bibitem{Mihajlovi2021LEAPLA}
Marko Mihajlovic, Yan Zhang, Michael~J Black, and Siyu Tang.
\newblock Leap: Learning articulated occupancy of people.
\newblock In {\em CVPR}, 2021.

\bibitem{Neverova2018}
Natalia Neverova, Riza~Alp G\"{u}ler, and Iasonas Kokkinos.
\newblock Dense pose transfer.
\newblock {\em ECCV}, 2018.

\bibitem{Niemeyer2021GIRAFFERS}
Michael Niemeyer and Andreas Geiger.
\newblock Giraffe: Representing scenes as compositional generative neural feature fields.
\newblock {\em CVPR}, pages 11448--11459, 2021.

\bibitem{narf}
Atsuhiro Noguchi, Xiao Sun, Stephen Lin, and Tatsuya Harada.
\newblock Neural articulated radiance field.
\newblock In {\em IEEE/CVF ICCV}, 2021.

\bibitem{OrEl2021StyleSDFH3}
Roy Or-El, Xuan Luo, Mengyi Shan, Eli Shechtman, Jeong~Joon Park, and Ira Kemelmacher-Shlizerman.
\newblock Stylesdf: High-resolution 3d-consistent image and geometry generation.
\newblock {\em ArXiv}, abs/2112.11427, 2021.

\bibitem{Palafox2021NPMsNP}
Pablo Palafox, Alja{\v{z}} Bo{\v{z}}i{\v{c}}, Justus Thies, Matthias Nie{\ss}ner, and Angela Dai.
\newblock Npms: Neural parametric models for 3d deformable shapes.
\newblock In {\em IEEE/CVF ICCV}, 2021.

\bibitem{park2019deepsdf}
Jeong~Joon Park, Peter Florence, Julian Straub, Richard Newcombe, and Steven Lovegrove.
\newblock Deepsdf: Learning continuous signed distance functions for shape representation.
\newblock {\em CVPR}, 2019.

\bibitem{Park2019DeepSDFLC}
Jeong~Joon Park, Peter~R. Florence, Julian Straub, Richard~A. Newcombe, and S. Lovegrove.
\newblock Deepsdf: Learning continuous signed distance functions for shape representation.
\newblock {\em 2019 (CVPR)}, pages 165--174, 2019.

\bibitem{Peng2021AnimatableNR}
Sida Peng, Junting Dong, Qianqian Wang, Shangzhan Zhang, Qing Shuai, Xiaowei Zhou, and Hujun Bao.
\newblock Animatable neural radiance fields for modeling dynamic human bodies.
\newblock In {\em ICCV}, 2021.

\bibitem{neuralbody}
Sida Peng, Yuanqing Zhang, Yinghao Xu, Qianqian Wang, Qing Shuai, Hujun Bao, and Xiaowei Zhou.
\newblock Neural body: Implicit neural representations with structured latent codes for novel view synthesis of dynamic humans.
\newblock {\em CVPR}, 2021.

\bibitem{smplpix}
Sergey Prokudin, Michael~J. Black, and Javier Romero.
\newblock Smplpix: Neural avatars from 3d human models.
\newblock {\em WACV}, 2021.

\bibitem{Pumarola_2018_CVPR}
Albert Pumarola, Antonio Agudo, Alberto Sanfeliu, and Francesc Moreno-Noguer.
\newblock Unsupervised person image synthesis in arbitrary poses.
\newblock In {\em CVPR}, June 2018.

\bibitem{anr}
Amit Raj, Julian Tanke, James Hays, Minh Vo, Carsten Stoll, and Christoph Lassner.
\newblock Anr: Articulated neural rendering for virtual avatars.
\newblock {\em CVPR}, pages 3721--3730, 2021.

\bibitem{Remelli2022DrivableVA}
Edoardo Remelli, Timur~M. Bagautdinov, Shunsuke Saito, Chenglei Wu, Tomas Simon, Shih-En Wei, Kaiwen Guo, Zhe Cao, Fabi{\'a}n Prada, Jason~M. Saragih, and Yaser Sheikh.
\newblock Drivable volumetric avatars using texel-aligned features.
\newblock {\em ACM SIGGRAPH}, 2022.

\bibitem{Saito2019PIFuPI}
Shunsuke Saito, Zeng Huang, Ryota Natsume, Shigeo Morishima, Angjoo Kanazawa, and Hao Li.
\newblock Pifu: Pixel-aligned implicit function for high-resolution clothed human digitization.
\newblock {\em IEEE/CVF ICCV}, pages 2304--2314, 2019.

\bibitem{Saito2020PIFuHDMP}
Shunsuke Saito, Tomas Simon, Jason~M. Saragih, and Hanbyul Joo.
\newblock Pifuhd: Multi-level pixel-aligned implicit function for high-resolution 3d human digitization.
\newblock {\em 2020 (CVPR)}, pages 81--90, 2020.

\bibitem{Saito2021SCANimateWS}
Shunsuke Saito, Jinlong Yang, Qianli Ma, and Michael~J. Black.
\newblock Scanimate: Weakly supervised learning of skinned clothed avatar networks.
\newblock {\em 2021 (CVPR)}, pages 2885--2896, 2021.

\bibitem{feanet}
Kripasindhu Sarkar, Dushyant Mehta, Weipeng Xu, Vladislav Golyanik, and Christian Theobalt.
\newblock Neural re-rendering of humans from a single image.
\newblock In {\em ECCV}, 2020.

\bibitem{SiaroSLS2017}
Aliaksandr Siarohin, Enver Sangineto, Stephane Lathuiliere, and Nicu Sebe.
\newblock Deformable {GANs} for pose-based human image generation.
\newblock In {\em CVPR}, 2018.

\bibitem{vgg}
K. Simonyan and Andrew Zisserman.
\newblock Very deep convolutional networks for large-scale image recognition.
\newblock {\em CoRR}, abs/1409.1556, 2015.

\bibitem{anerf}
Shih-Yang Su, Frank Yu, Michael Zollhoefer, and Helge Rhodin.
\newblock A-nerf: Articulated neural radiance fields for learning human shape, appearance, and pose.
\newblock In {\em NeurIPS}, 2021.

\bibitem{Tewari2020StateOT}
Ayush Tewari, Ohad Fried, Justus Thies, Vincent Sitzmann, Stephen Lombardi, Kalyan Sunkavalli, Ricardo Martin-Brualla, Tomas Simon, Jason~M. Saragih, Matthias Nie{\ss}ner, Rohit Pandey, S. Fanello, Gordon Wetzstein, Jun-Yan Zhu, Christian Theobalt, Maneesh Agrawala, Eli Shechtman, Dan~B. Goldman, and Michael Zollhofer.
\newblock State of the art on neural rendering.
\newblock {\em Computer Graphics Forum}, 2020.

\bibitem{dnr}
Justus Thies, Michael Zollh\"{o}fer, and Matthias Nießner.
\newblock Deferred neural rendering: image synthesis using neural textures.
\newblock {\em ACM Transactions on Graphics (TOG)}, 38, 2019.

\bibitem{Tiwari2021NeuralGIFNG}
Garvita Tiwari, Nikolaos Sarafianos, Tony Tung, and Gerard Pons-Moll.
\newblock Neural-gif: Neural generalized implicit functions for animating people in clothing.
\newblock In {\em ICCV}, 2021.

\bibitem{surreal}
G. Varol, J. Romero, X. Martin, Naureen Mahmood, Michael~J. Black, I. Laptev, and C. Schmid.
\newblock Learning from synthetic humans.
\newblock {\em CVPR}, pages 4627--4635, 2017.

\bibitem{Wang2021MetaAvatarLA}
Shaofei Wang, Marko Mihajlovic, Qianli Ma, Andreas Geiger, and Siyu Tang.
\newblock Metaavatar: Learning animatable clothed human models from few depth images.
\newblock {\em NeurIPS}, 2021.

\bibitem{ARAH:2022:ECCV}
Shaofei Wang, Katja Schwarz, Andreas Geiger, and Siyu Tang.
\newblock Arah: Animatable volume rendering of articulated human sdfs.
\newblock In {\em European Conference on Computer Vision}, 2022.

\bibitem{vid2vid}
Ting-Chun Wang, Ming-Yu Liu, Jun-Yan Zhu, Guilin Liu, Andrew Tao, Jan Kautz, and Bryan Catanzaro.
\newblock Video-to-video synthesis.
\newblock In {\em NeurIPS}, 2018.

\bibitem{pix2pixhd}
Ting-Chun Wang, Ming-Yu Liu, Jun-Yan Zhu, Andrew Tao, Jan Kautz, and Bryan Catanzaro.
\newblock High-resolution image synthesis and semantic manipulation with conditional gans.
\newblock In {\em CVPR}, 2018.

\bibitem{ssim}
Zhou Wang, A. Bovik, H.~R. Sheikh, and E.~P. Simoncelli.
\newblock Image quality assessment: from error visibility to structural similarity.
\newblock {\em IEEE Transactions on Image Processing}, 13:600--612, 2004.

\bibitem{Weng2022HumanNeRFFR}
Chung-Yi Weng, Brian Curless, Pratul~P. Srinivasan, Jonathan~T. Barron, and Ira Kemelmacher-Shlizerman.
\newblock Humannerf: Free-viewpoint rendering of moving people from monocular video.
\newblock {\em ArXiv}, abs/2201.04127, 2022.

\bibitem{Xu2011VideobasedCC}
Feng Xu, Yebin Liu, Carsten Stoll, James Tompkin, Gaurav Bharaj, Qionghai Dai, Hans-Peter Seidel, Jan Kautz, and Christian Theobalt.
\newblock Video-based characters: creating new human performances from a multi-view video database.
\newblock {\em ACM SIGGRAPH}, 2011.

\bibitem{Yoon2022LearningMA}
Jae~Shin Yoon, Duygu Ceylan, Tuanfeng~Y. Wang, Jingwan Lu, Jimei Yang, Zhixin Shu, and Hyunjung Park.
\newblock Learning motion-dependent appearance for high-fidelity rendering of dynamic humans from a single camera.
\newblock {\em 2022 IEEE/CVF Conference on Computer Vision and Pattern Recognition (CVPR)}, pages 3397--3407, 2022.

\bibitem{lpip}
Richard Zhang, Phillip Isola, Alexei~A. Efros, E. Shechtman, and O. Wang.
\newblock The unreasonable effectiveness of deep features as a perceptual metric.
\newblock {\em CVPR}, pages 586--595, 2018.

\bibitem{Zheng2021DeepMultiCapPC}
Yang Zheng, Ruizhi Shao, Yuxiang Zhang, Tao Yu, Zerong Zheng, Qionghai Dai, and Yebin Liu.
\newblock Deepmulticap: Performance capture of multiple characters using sparse multiview cameras.
\newblock In {\em Proceedings of the IEEE/CVF International Conference on Computer Vision}, pages 6239--6249, 2021.

\bibitem{Zheng2021PaMIRPM}
Zerong Zheng, Tao Yu, Yebin Liu, and Qionghai Dai.
\newblock Pamir: Parametric model-conditioned implicit representation for image-based human reconstruction.
\newblock {\em TPAMI}, PP, 2021.

\bibitem{Zhou2021CIPS3DA3}
Peng Zhou, Lingxi Xie, Bingbing Ni, and Qi Tian.
\newblock Cips-3d: A 3d-aware generator of gans based on conditionally-independent pixel synthesis.
\newblock {\em ArXiv}, 2021.

\bibitem{zhu2019progressive}
Zhen Zhu, Tengteng Huang, Baoguang Shi, Miao Yu, Bofei Wang, and Xiang Bai.
\newblock Progressive pose attention transfer for person image generation.
\newblock In {\em CVPR}, pages 2347--2356, 2019.

\end{thebibliography}
}

\end{document}